\documentclass[sn-mathphys,iicol]{sn-jnl}

\graphicspath{{./figures/}}

\usepackage{mdframed}
\usepackage{multicol}
\usepackage{epsfig}
\usepackage{epstopdf}
\newcommand{\comment}[1]{}

\jyear{2022}%

\theoremstyle{thmstyleone}%
\theoremstyle{thmstyletwo}%

\theoremstyle{thmstylethree}%

\begin{document}

\title[\hspace{75mm} State-of-the-Art Review of Design of Experiments for PINN]{State-of-the-Art Review of Design of Experiments for Physics-Informed Deep Learning}


\author*[1]{\fnm{Sourav} \sur{Das}}\email{sds2019@mail.ubc.ca}

\author[1]{\fnm{Solomon} \sur{Tesfamariam}}\email{solomon.tesfamariam@ubc.ca}

\affil*[1]{\orgdiv{School of Engineering}, \orgname{The University of British Columbia, Okanagan Campus}, \orgaddress{\street{3333 University Way}, \city{Kelowna}, \postcode{V1V1V7}, \state{BC}, \country{Canada}}}

\abstract{This paper presents a comprehensive review of the design of experiments used in the surrogate models. In particular, this study demonstrates the necessity of the design of experiment schemes for the Physics-Informed Neural Network (PINN), which belongs to the supervised learning class. Many complex partial differential equations (PDEs) do not have any analytical solution; only numerical methods are used to solve the equations, which is computationally expensive. In recent decades, PINN has gained popularity as a replacement for numerical methods to reduce the computational budget. PINN uses physical information in the form of differential equations to enhance the performance of the neural networks. Though it works efficiently, the choice of the design of experiment scheme is important as the accuracy of the predicted responses using PINN depends on the training data. In this study, five different PDEs are used for numerical purposes, i.e., viscous Burger's equation, Shr\"{o}dinger equation, heat equation, Allen-Cahn equation, and Korteweg-de Vries equation. A comparative study is performed to establish the necessity of the selection of a DoE scheme. It is seen that the Hammersley sampling-based PINN performs better than other DoE sample strategies.}


\maketitle

\comment{
\begin{figure*}
    \centering
\begin{mdframed}
    \begin{multicols}{2}
        \textbf{Nomenclature}
        \begin{description}
            \item[ANN] Artificial Neural Network
            \item[BBD] Box-Behnken Design
            \item[BLUP] Best Linear Unbiased Prediction
            \item[CCD] Central Composite Design
            \item[CCC] CCD - Circumscribed
            \item[CCI] CCD - Inscribed
            \item[CCF] CCD - Faced
            \item[CVT] Centroidal Voronoi Tesselation
            \item[DoE] Design of Experiments
            \item[FEM] Finite Element Method
            \item[HDMR] High Dimensional Model Representation
            \item[LHS] Latin Hypercube Sampling
            \item[MARS] Multivariate Adaptive Regression Spline
            \item[MCMC] Markov Chain Monte Carlo
            \item[MCS] Monte Carlo Simulation
            \item[ML] Machine Learning
            \item[MSE] Mean Square Error
            \item[OA] Orthogonal Arrays
            \item[PCE] Polynomial Chaos Expansion
            \item[PINN] Physics-Informed Neural Network
            \item[RBF] Radial Basis Function
            \item[$\mathcal{SG}$] Sparse Grid
            \item[SVR] Support Vector Regression
        \end{description}
    \end{multicols}
\end{mdframed}
\end{figure*}
}

\section{Introduction}

\begin{figure*}[h!]
    \centering
    \includegraphics[width=0.8\textwidth]{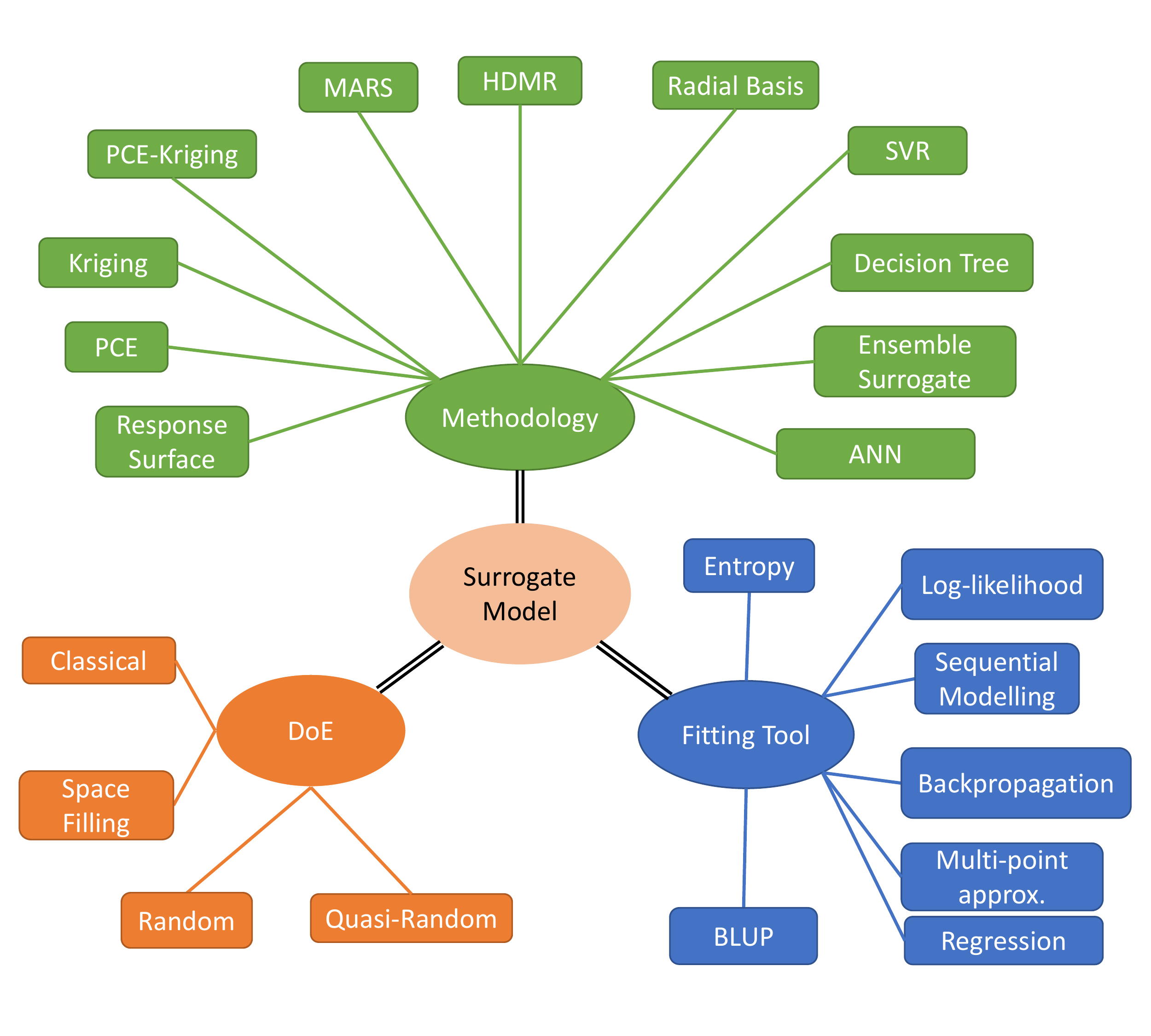}
    \caption{A schematic diagram shows the commonly used surrogate models}
    \label{fig:surr1}
\end{figure*}

Quantifying uncertainties in a physical system is an active area of research in the science and engineering fields \cite{der2009aleatory}. In design optimization under uncertainty, Monte Carlo simulation (MCS) \cite{sheppard1969computer} is the oldest and most widely used technique. The MCS, however, requires large number of model evaluations and computationally is costly for high-fidelity models. Surrogate-assisted models, developed using Design of Experiment (DoE) (Fig.~\ref{fig:surr1}), are attractive options.

Once the training data-set through the DoE are generated, different methodologies are used to build the surrogate model (Fig.~\ref{fig:surr1}), e.g. response surface method \cite{montgomery2017design}, stochastic response surface method \cite{rathi2019sequential}, polynomial chaos expansion (PCE) \cite{xiu2002wiener,sudret2008global}, Kriging \cite{matheron1967kriging,das2020optimization}, PCE-Kriging \cite{schobi2015polynomial}, high dimensional model representation (HDMR) \cite{rathi2021development,rathi2021improved}, radial basis function (RBF) \cite{wang2006radial,de2013new}, support vector regression (SVR) \cite{xiao2015efficient}, artificial neural network \cite{hassoun1995fundamentals}, ensemble surrogate \cite{das2021optimal}, etc. The surrogate models minimize the noise between actual and predicted responses using fitting tools (shown in Fig.~\ref{fig:surr1}), based on the knowledge of DoE samples. The purpose of using DoE is that the samples should cover the entire design space; otherwise, predicted responses will be erroneous. Fig.~\ref{fig:surr1} illustrates a schematic diagram of the commonly used surrogate models, DoE schemes and the fitting tools.

Machine learning is a widely used methodology to develop forecasting models and support decision making under uncertainty \citep{murphy2012machine, mitchell1997machine}. The recent research trends devotes on the machine learning techniques that can be used as a surrogate model for regression analysis. Feed-forward neural networks, which is a special class of supervised machine learning technique, a data-driven network, is used in this paper. In this study, we are focusing only on physics-informed neural networks (PINNs), where physical conditions are imposed on feed-forward neural network. 
The PINN has been widely used in various fields. An overview of variants and applications of PINN is presented here. Raissi and Karniadakis \cite{raissi2018hidden} proposed PINN to identify parameters in partial differential equations (PDEs) like the Kortewegde Vries Equation and the Navier-Stokes Equation \citep{raissi2020hidden}. The solution to the alloy solidification benchmark problem was investigated by Rad \textit{et al.} \cite{rad2020theory}. Shin \textit{et al.} \cite{shin2020convergence} showed the convergence of PINN to solve PDEs towards the theoretical solution of PDEs using the Schrader approach. PINN is also used to solve physical problems by replacing the finite element method (FEM). Haghighat \textit{et al.} \cite{haghighat2020deep} used PINN to solve the linear elastic problems in the field of solid mechanics. The direct and inverse heat conduction problems of materials were studied by He \textit{et al.} \cite{he2021physics}. Olivieri \textit{et al.} \cite{olivieri2021physics} proposed a PINN based methodology for near-field acoustic holography for contact-less analysis of vibrating structures. PINN is also used to forecast the remaining useful life (RUL) of the rotating machinery in wind firms. Eker \textit{et al.} \cite{eker2019new} proposed a physics-informed model for short-term prediction of the RUL of the rotating machinery. Yucesan and Viana \cite{yucesan2020physics} studied wind turbine main bearing fatigue using PINN. Its application is extended into fluid-structure interaction problems. The performance of vortex and wake induced vibration of the cylinder is studied by Cheng \textit{et al.} \cite{cheng2021deep} using PINN based on Reynolds Average Navier Stokes equations. Bai and Zhang \cite{bai2021machine} extended the above work from laminar flow to turbulent flow by adding viscosity into the equations using PINN. To forecast the flow field without any simulation, Sun \textit{et al.} \cite{sun2020surrogate} proposed a PINN-based framework which includes the physical equations, initial and boundary conditions. Tang \textit{et al.} \cite{tang2021transfer} proposed a transfer learning based PINN to enhance learning efficiency for vortex induced vibration. Another application of the civil engineering field is seismic wave propagation, where time domain wave equations are computationally expensive as they need a lot of memory to store the wavefield solutions. With this in view, Song \textit{et al.} \cite{song2022versatile} proposed a PINN-based framework to solve the Helmholtz equation for obtaining wavefield solutions. Karimpouli and Tahmasebi \cite{karimpouli2020physics}, Alkhalifah \textit{et al.} \cite{alkhalifah2020wavefield} demonstrated the efficacy of PINN in modelling the wave equation in the time- and frequency-domain. The solutions of isotropic and anisotropic $P$-wave eikonal equations using PINN were shown by Smith \textit{et al.} \cite{smith2020eikonet}, Waheed \textit{et al.} \cite{waheed2020anisotropic}. To solve the scattered pressure wavefield, Alkhalifah \textit{et al.} \cite{alkhalifah2020wavefield} proposed a PINN-based framework. The other applications of PINN are discussed briefly. Patel \textit{et al.} \cite{patel2022thermodynamically} investigated thermodynamically consistent PINN for hyperbolic shock hydrodynamics systems. The forward and inverse problems related to the nonlinear diffusivity and Biot's equation were investigated by Kadeethum \textit{et al.} \cite{kadeethum2020physics}. They also studied the effects of noisy measurements on inverse problems and addressed the challenge of selecting the hyperparameters of the inverse model. Lu \textit{et al.} \cite{lu2021physics} proposed a deep learning algorithm for topology optimization that includes PINN and hard constraints. Schiassi \textit{et al.} \cite{schiassi2021physics} proposed a novel technique of PINN for the solution of point kinetics equations with temperature feedback for nuclear reactor dynamics. They developed a framework that is a combination of PINN with the Theory of Functional Connections and Extreme Learning Machines. The corrosion fatigue modeling, consisting of crack growth and damage bias due to corrosion of the aircraft wings, was proposed by Dourado and Viana \cite{dourado2019physics} using PINN. The multifidelity simulation technique, which is a combination of low-fidelity and high-fidelity simulations using PINN, was studied by Penwarden \textit{et al.} \cite{penwarden2021multifidelity} to reduce the computational cost. Kovacs \textit{et al.} \cite{kovacs2022magnetostatics} used PINN to solve magnetostatic and micromagnetic problems. The design of electromagnetic metamaterials using PINN was proposed by Fang and Zhan \cite{fang2019deep}. Aside from the applications of PINN in various fields, various variants of PINN have been developed, such as conditional PINN \citep{kovacs2022conditional}, conservative PINN (cPINN) \citep{jagtap2020conservative}, variational PINN (vPINN) \citep{kharazmi2019variational}, parareal PINN (pPINN) \citep{meng2020ppinn}, stochastic PINN (sPINN) \citep{o2021stochastic}, fractional PINN (fPINN) \citep{pang2019fpinns}, nonlocal PINN (nPINN) \citep{pang2020npinns}, extended PINN (xPINN) \citep{jagtap2020extended} etc. Shukla \textit{et al.} \cite{SHUKLA2021110683} proposed a domain decomposition based parallel PINN where all hyperparameters of each neural network are optimized separately in each sub-domain. Kharazmi \textit{et al.} \cite{kharazmi2021hp} proposed a novel \textit{hp}-Variational PINN based on the sub-domain Petrov–Galerkin method where domain decomposition was used as \textit{h}-refinement and projection onto space of high order polynomials was used as \textit{p}-refinement.

The outline of the article is structured as follows. Section~\ref{sec:DOE} describes an overview of DoE schemes. Section~\ref{sec:ML} provides a comprehensive summary of machine learning techniques in general and PINN.  Different applications of PINN using the DoE schemes are presented in Section~\ref{sec:numerical}. Finally, Section~\ref{sec:conclude} presents the concluding remarks.

\section{Design of Experiment Sampling Strategy}\label{sec:DOE}

\begin{figure*}[h!]
    \centering
    \includegraphics[width=1.0\textwidth]{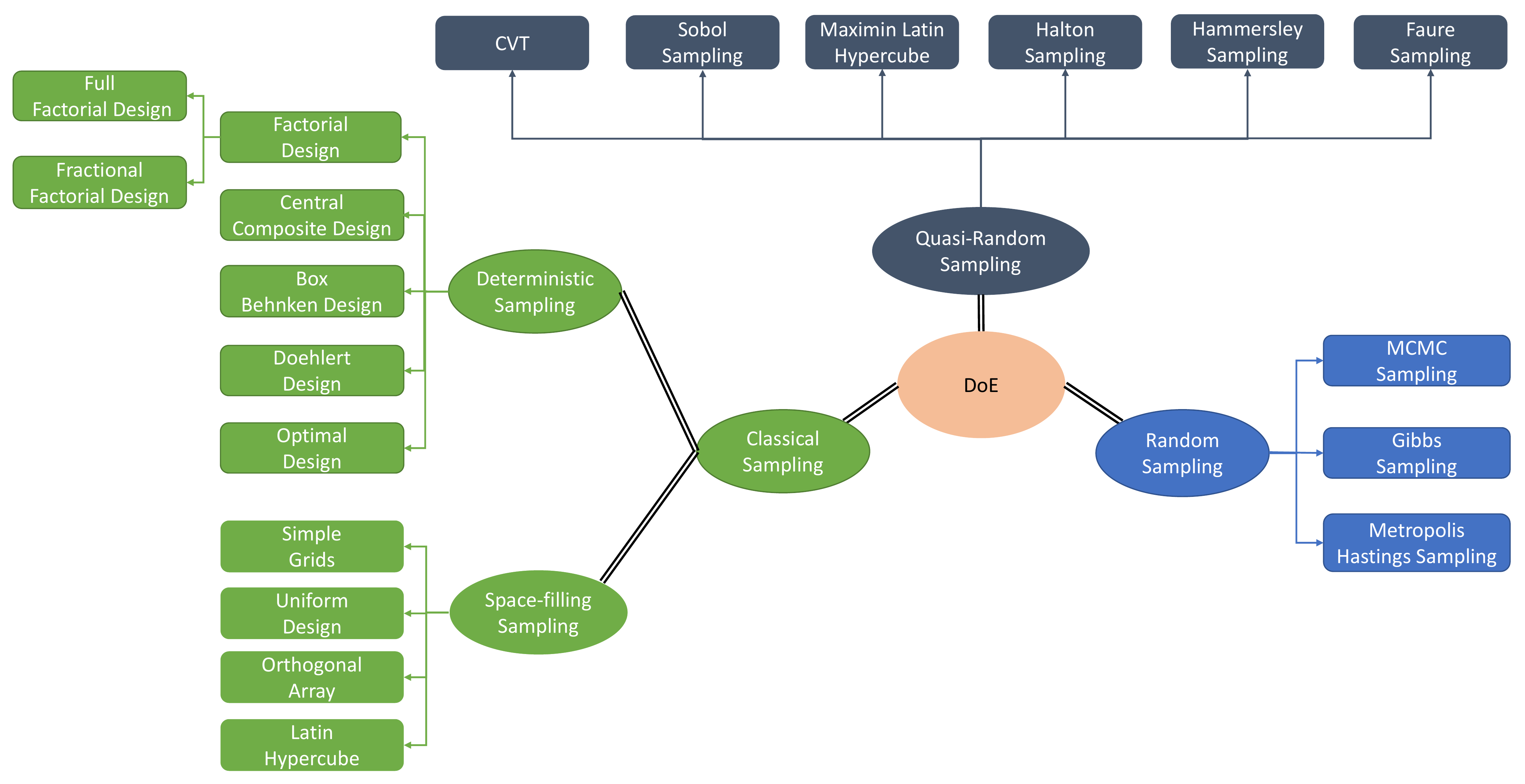}
    \caption{Commonly used DoE sampling strategies}
    \label{fig:DOE}
\end{figure*}

Experiments used to be performed by trial-and-error to find sample (or training samples or support) points \citep{montgomery2017design}. Fisher \cite{fisher1992arrangement} introduced the concept of a design of experiment (DoE) strategy in the early 20th century for investigating the probabilistic behaviour of agricultural crop systems. The DoEs are used in the engineering field \citep{koehler19969,santner2003design} and are gaining popularity due to the reduction in computational costs. The DoE can be grouped into two categories, i.e., Classical DoE \citep{anderson2018design,lorenzen1993design,mason2003statistical,antony2014design,montgomery2017design} and Modern DoE \citep{sacks1989designs,chen2002review,crary2002design,santner2003design}. The DoE samples  are generated using two approaches, i.e., domain-based (or non-adaptive or model-free) and response-based (or adaptive or model-based or sequential) approaches. In the domain-based approach, the DoE samples are generated based on the information obtained from the design space, whereas in the response-based approach, the samples are chosen from the information obtained from the surrogate models. In this paper, extensive reviews on generation of DoE sample strategy are presented. The study is mainly focused on four different sampling schemes, i.e., deterministic sampling (classical DoE), space-filling sampling, random sampling, and quasi-random sampling. The DoE schemes are shown in Fig.~\ref{fig:DOE} and further discussed below.\ 

In the deterministic scheme, the sample points are generated uniformly over the design space, and the number of sample points is also fixed with the aim of minimizing the random error of the simulation. Therefore, the samples are generated in such a manner that most of the samples are located at the boundary of the design space, with few samples in the interior space. There are various type of classical DoE scheme available, factorial design \cite{montgomery2017design}, central composite design \cite{box1951wilson}, Box-Behnken design \cite{ferreira2007box}, optimal design \cite{kiefer1959optimum}, etc.\
Space-filling sampling also lies within the classical DoE approach. This sampling technique is mainly used to enhance accuracy of the surrogate models. Initially, a surrogate model is built with few samples, and that model is updated iteratively by adding a sample to the previous DoE set used until a satisfactory accuracy is achieved. In this category, simple grids, Latin hypercube \cite{helton2003latin}, orthogonal array \cite{hedayat1999orthogonal}, and uniform design \cite{li2002model} are the most popular DoE generation schemes. \

Random sampling and quasi-random sampling lie within the modern DoE approach. In random sampling, Markov chain Monte Carlo (MCMC) sampling, \cite{kroese2013handbook}, Gibbs sampling \cite{gelfand2000gibbs}, and Metropolis-Hastings sampling \cite{chib1995understanding} are popular. The samples are generated with an equally likely probability of occurrence of samples within the design space. This technique is suitable only when the proper knowledge of design space is available because the clustering or poor coverage regions of samples have been observed for high-dimensional problems. In quasi-random sampling, the popular schemes are centroidal Voronoi tessellation (CVT) \cite{du1999centroidal}, maximin Latin hypercube \cite{johnson1990minimax}, Sobol sampling \cite{sobol1967distribution}, Halton sampling \cite{halton1960efficiency}, Hammersley sampling \cite{wong1997sampling}, Faure sampling \cite{faure1982discrepance}, etc.


\subsection{Full \& Fractional Factorial Design}

\begin{figure*}[h!]
\centering     
\subfigure[]{\label{fig:2d_lev1_full}\includegraphics[width=0.49\textwidth]{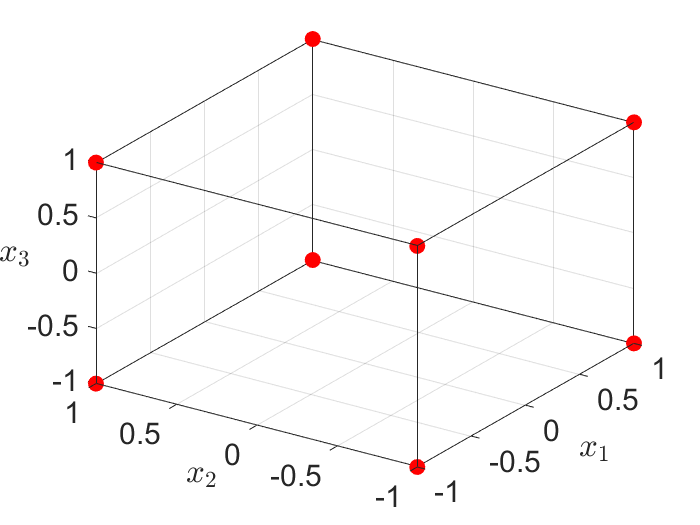}} 
\subfigure[]{\label{fig:2d_lev2_fract}\includegraphics[width=0.49\textwidth]{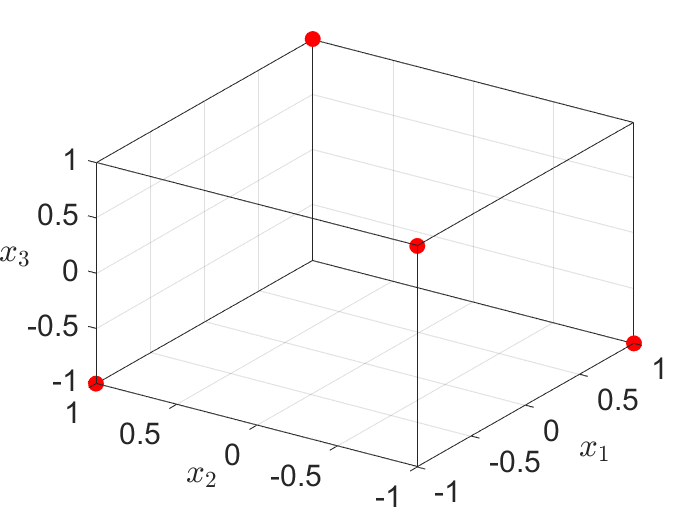}}
\caption{2-levels \protect\subref{fig:2d_lev1_full} full and  \protect\subref{fig:2d_lev2_fract} fractional factorial design}
\label{fig:fact_des}
\end{figure*}

Factorial design is one of the oldest techniques for the design of experiment samples \cite{montgomery2017design}. Every input variable is discretized into two or three levels. Fig.~\ref{fig:2d_lev1_full} shows a two-level full factorial design. Every input variable is bounded by lower and upper limits, which are the minimum and maximum values of the variable, expressed by (-1) and (+1). For three-level factorial design, a sample point is taken at the center of the variable, which is the mean value of the input variable. More concisely, a two-level design takes sample points at the lower and upper limits of the variable, whereas a three-level design takes sample points at the lower, upper, and center. In general, the number of runs or simulations required is expressed as $2^n$ or $3^n$ for two- and three-level designs, respectively, where $n$ is the number of input variables. Fig.~\ref{fig:2d_lev1_full} shows a $2^3$ full factorial design.

Though full factorial design is easy to use, it has some drawbacks as the number of runs or simulations depends on the number of input variables considered. For high dimensional problems, the number of runs increases with an increasing number of variables, which becomes computationally expensive for high-fidelity models. The fractional factorial design lowers the computational cost where fewer sampling points are generated than in the full factorial design. For fractional factorial design, the number of samples is expressed as $2^{n-k}$ and $3^{n-k}$, for two- and three-level design, respectively. Fig.~\ref{fig:fact_des} depicts a $2^3$ design, i.e. two-level and three factors for full and fractional factorial design.
Though factorial design is a commonly used DoE scheme, it fails to capture the nonlinearity of the system responses as it considers only the lower, center, and upper values of the variables. 

\subsection{Central Composite Design}

\begin{figure*}[h!]
\centering     
\subfigure[]{\label{fig:ccd_circum}\includegraphics[width=0.49\textwidth]{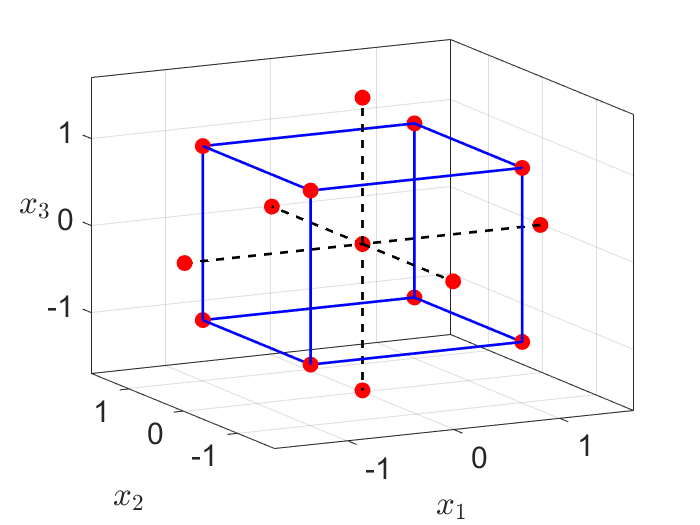}} 
\subfigure[]{\label{fig:ccd_ins}\includegraphics[width=0.49\textwidth]{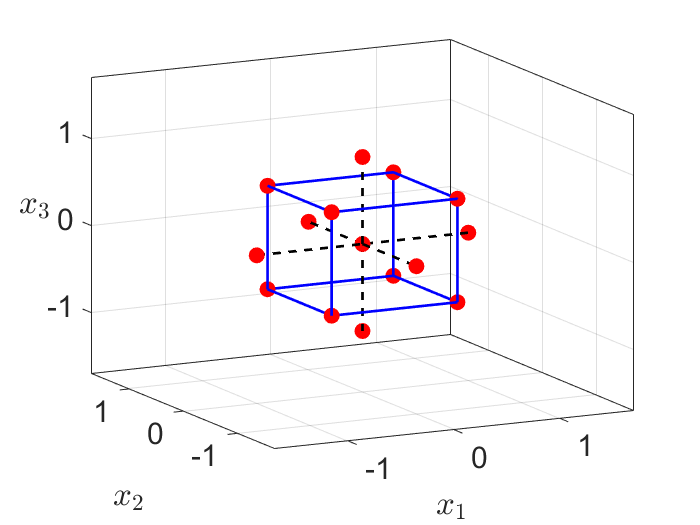}}
\subfigure[]{\label{fig:ccd_faced}\includegraphics[width=0.49\textwidth]{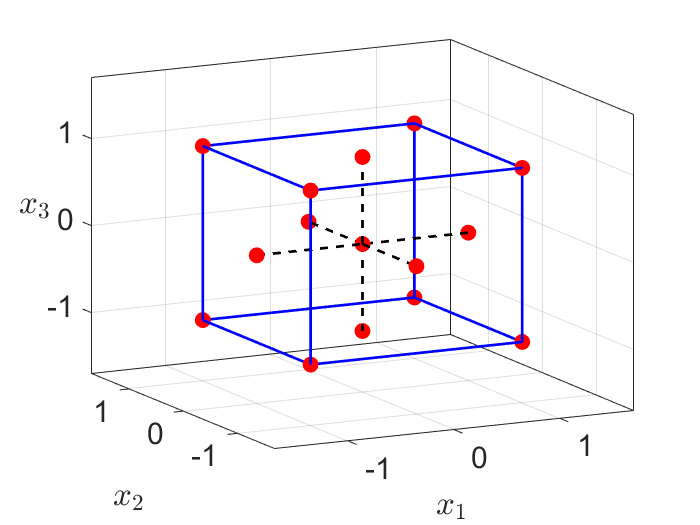}}
\caption{Central composite design; \protect\subref{fig:ccd_circum} Circumscribed, \protect\subref{fig:ccd_ins} Inscribed and \protect\subref{fig:ccd_faced} Faced}
\label{fig:ccd}
\end{figure*}

The central composite design (CCD), proposed by Box and Wilson \cite{box1951wilson}, is a variant of factorial design where additional sampling points are considered at the center and axial directions. Fig.~\ref{fig:ccd} shows the two-level CCD for three factors. The extremity samples are located at the lower and upper limits of each variable, which is the same as in full factorial design. Additionally, two axial points for each variable and the center point of the hypercube are considered. For a two-level central composite design, the number of simulations required for a given number of input variables is $2n + 2n +1$. The CCD are categorized into three types, i.e., circumscribed (CCC), inscribed (CCI) and faced (CCF). In the CCC design, the axial points are located at the outside of the cube where the cube points take (-1) and (+1) values. CCI design is the same as CCC design, but scaled where the axial points are taken (-1) and (+1) values and cube points are located inside the cube. Unlike in CCC design, where axial points are located on the outside of the cube, in CCF design, the axial points are placed on the face of the cube. Fig.~\ref{fig:ccd} shows all three types of CCD for three factors.

\subsection{Box-Behnken Design}

\begin{figure}[h!]
    \centering
    \includegraphics[width=0.49\textwidth]{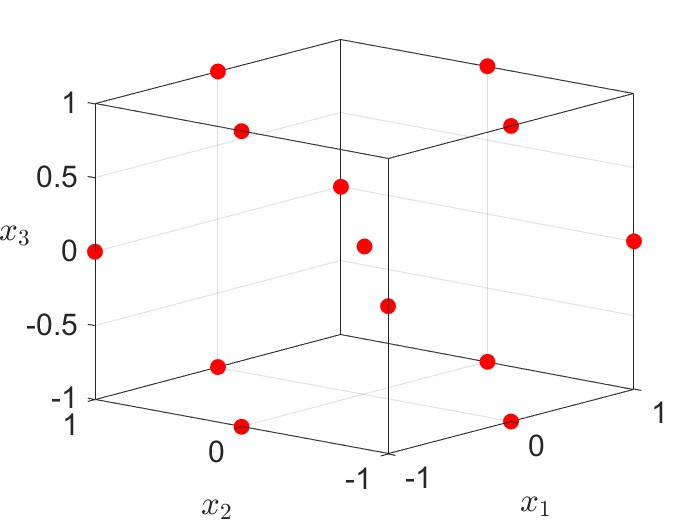}
    \caption{ Box-Behnken design}
    \label{fig:bbd}
\end{figure}

The Box-Behnken design (BBD) \cite{ferreira2007box} is similar to CCD, where the sample points are located at the midpoint of each edge and one at the center of the hypercube. It does not have any corners or extreme points. Fig.~\ref{fig:bbd} illustrates the Box-Behnken design for three factors. A comparison of DoE samples obtained from all types of CCD and BBD is shown in Table~\ref{tab:comp2} for three factors. It is seen that BBD requires less number of simulations compared to other CCDs. Like factorial design and CCD, it also suffers ``curse of dimensionality" for high dimensional problems as the number of simulations increases with the increasing number of input variables, which affect the computational cost.

\subsection{Doehlert Design}

\begin{figure}[h!]
\centering     
\subfigure[]{\label{fig:DD1}\includegraphics[width=0.49\textwidth]{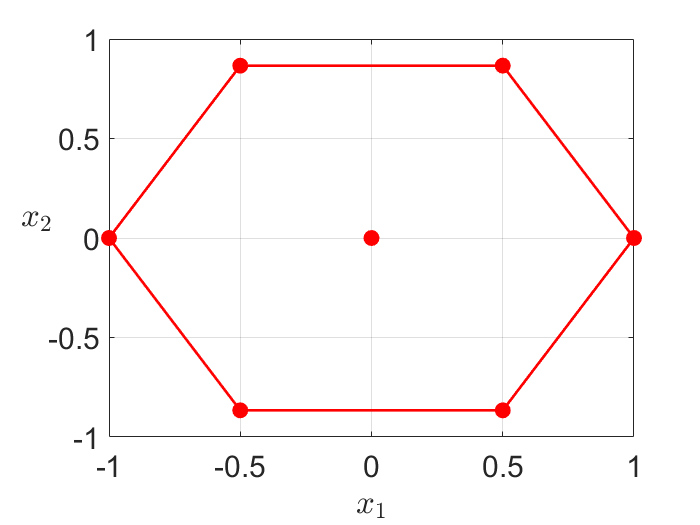}} 
\subfigure[]{\label{fig:DD2}\includegraphics[width=0.49\textwidth]{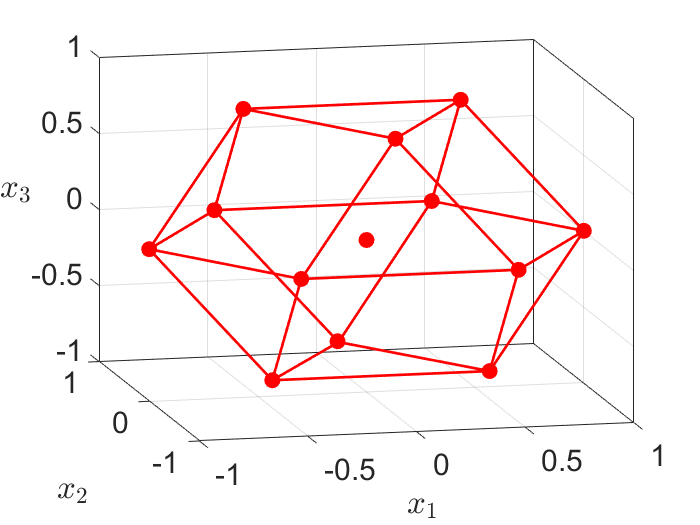}}
\caption{Doehlert design for \protect\subref{fig:DD1} two and \protect\subref{fig:DD2} three factors}
\label{fig:Doehlert}
\end{figure}

Doehlert \cite{doehlert1970uniform} proposed another efficient design of experiment scheme that takes fewer simulations than CCD and BBD. This strategy has been proven more economical and highly-efficient than other deterministic DoE sample strategies \citep{hibbert2012experimental}. The main difference between this design and CCD and BBD is that the samples generated from the Doehlert design are not rotatable due to the number of estimations for different input variables or factors. At the same time, this design is efficient in filling the space uniformly. Doehlert design provides different structure for the different number of factors such as, for two variables, samples are generated by forming a circular domain. For three factors, it becomes spherical domain and for four or more factors, it becomes hyperspherical domain \citep{bezerra2008response}. This design allows the total number of simulation as $n^2+n+1$, where $n$ is the number of factors. Fig.~\ref{fig:Doehlert} illustrates the Doehlert design for two and three factors. Table~\ref{tab:comp2} shows the comparison of number of samples generated using factorial, central composite, Box-Behnken and Doehlert design for different factors.

\begin{table*}[h!]
    \centering
     \caption{A comparison of number of samples generated using factorial, central composite, Box-Behnken and Doehlert design for different factors}
    \begin{tabular}{|c |c |c c c c|}
    \hline
       \multirow{2}{*}{No. of Factors}  & \multirow{2}{*}{No. of Coefficient}& \multicolumn{4}{c|}{No. of Runs/ Simulations}\\ \cline{3-6}  
       & & Factorial & CCD & BBD & Doehlert \\
       \hline
       2 & 6 & 9 & 9 & - & 7\\
       3 & 10 & 27 & 15 & 13 & 13\\
       4 & 15 & 81 & 25 & 25 & 21 \\
       5 & 21 & 243 & 43 & 41 & 31\\
       6 & 28 & 729 & 77 & 61 & 43\\
       7 & 36 & 2187 & 143 & 85 & 57 \\
       8 & 45 & 6561 & 273 & 113 & 73\\
         \hline
    \end{tabular}
    \label{tab:comp2}
\end{table*}

\subsection{Optimal Design}

The optimal design, proposed by Kiefer and Wolfowitz \cite{kiefer1959optimum}, is one of the common classical DoE strategy in which the optimality of any design is achieved by minimizing the statistical parameters, such as variance of the predicted responses. Later, a number of optimal criteria were developed by Elfving \cite{elfving1952optimum}, Kiefer \cite{kiefer1961optimum}, Bondar \cite{bondar1983universal}, named as $\phi$-optimality criteria. The most common $\phi$-optimality criteria are $D$, $G$, $A$, $E$, $I$, $L$, $c$, $\phi_p$ and integrated mean square error, etc. The advantages of optimal design are that it requires less number of samples than other classical DoE approaches because it involves optimization of the predicted responses. The applications of this strategy can be found in \cite{breiman1991discussion,pukelsheim1993optimal,parkinson1993general,jin2003efficient}.


\subsection{Orthogonal Arrays }
\begin{figure*}[h!]
\centering     
\subfigure[]{\label{fig:OA}\includegraphics[width=0.49\textwidth]{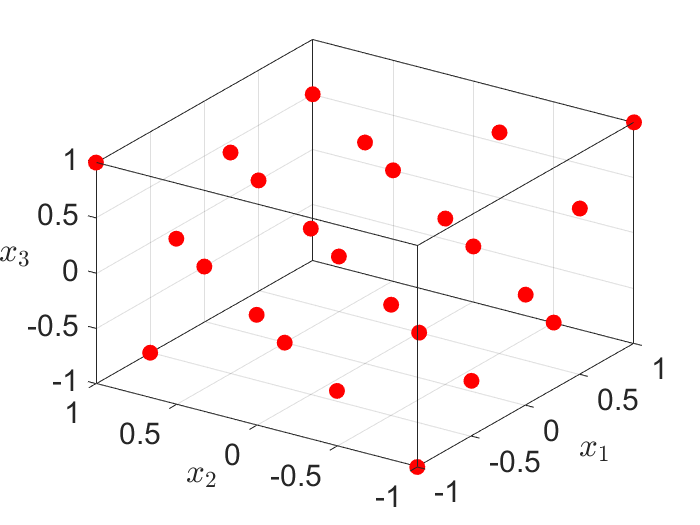}} 
\subfigure[]{\label{fig:OA_LHS}\includegraphics[width=0.49\textwidth]{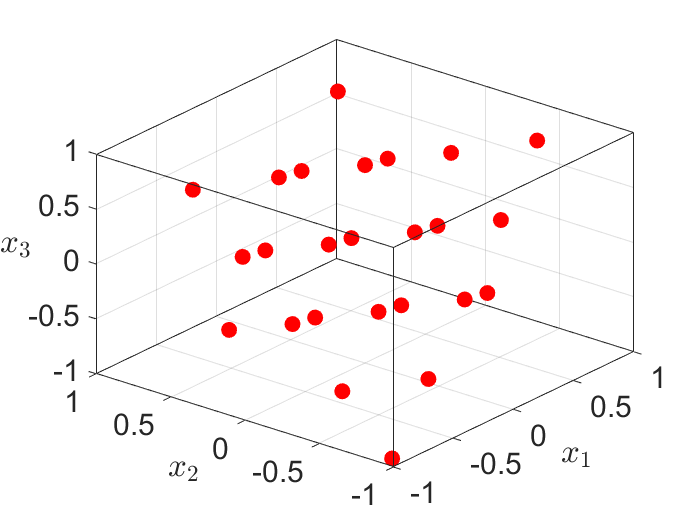}}
\caption{\protect\subref{fig:OA} Orthogonal array and \protect\subref{fig:OA_LHS} Orthogonal array-based Latin hypercube design}
\label{fig:ortho}
\end{figure*}

The orthogonal array, also known as Taguchi method \citep{taguchi1959linear,kacker1991taguchi,owen1992orthogonal,hedayat1999orthogonal}, is one of the most popular classical design of experiment schemes. An orthogonal array is expressed as $OA$ ($N, s_1^{m_1}, s_2^{m_2}, \ldots, s_p^{m_p}, \beta$) of strength $\beta$ is an $N \times m$ where $0\leq t \leq m$ and $m = m_1 + m_2 + \cdots + m_p$. Every column (i.e. $m_i$) has $s_i \geq 2$ elements such that for any strength $\beta$, all possible combinations of the elements appeared equally in the matrix. For example, $OA$(4,3,2,2) (sometimes expressed as $OA_4 (2^3)$) is the $4/2^3$ = 1/2 fraction of a $2^3$ full factorial design. Table~\ref{tab:comp3} displays $OA$(4,3,2,2) with the elements (0,0), (0,1), (1,0) and (1,1) appearing in the matrix only once. The orthogonal array for three factors is shown in Fig.~\ref{fig:ortho}.

\begin{table}[h!]
    \centering
     \caption{Orthogonal array, $OA(4,3,2,2)$}
    \begin{tabular}{|c| c c c|}
    \hline
       \multirow{2}{*}{Row No.}  &  \multicolumn{3}{c|}{Column No.}\\ \cline{2-4}  
       & 1 & 2 &3 \\
       \hline
      1& 0 & 0 & 0\\
     2 &  0&	1&	1\\
    3&	1&	0&	1\\
    4&	1&	1&	0\\
         \hline
    \end{tabular}
    \label{tab:comp3}
\end{table}

\subsection{Random Design}

\begin{figure*}[h!]
\centering     
\subfigure[]{\label{fig:gibbs}\includegraphics[width=0.49\textwidth]{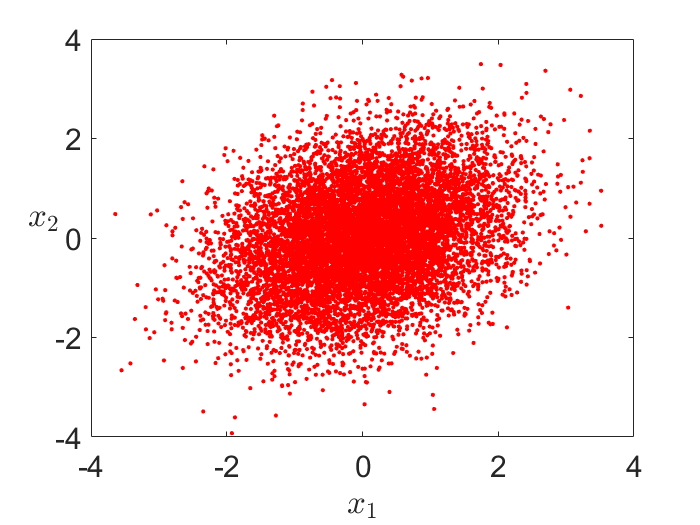}} 
\subfigure[]{\label{fig:mmh}\includegraphics[width=0.49\textwidth]{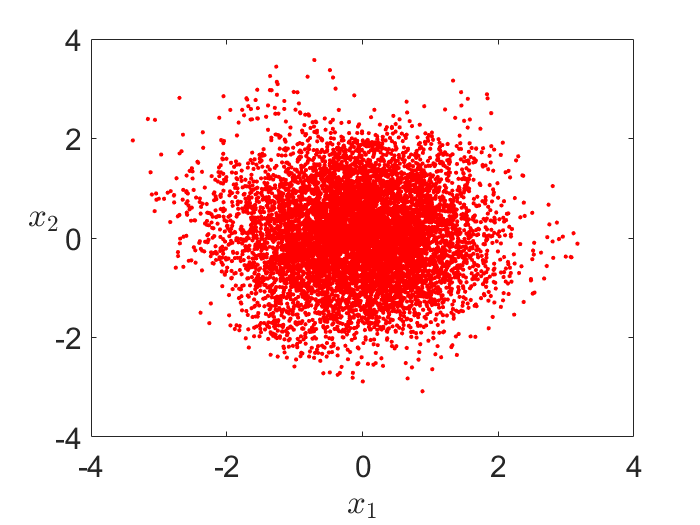}}
\caption{\protect\subref{fig:gibbs} Gibbs sampling and \protect\subref{fig:mmh} Metropolis-Hastings sampling}
\label{fig:random}
\end{figure*}

Random design, where the probability of occurrence of all samples is equal, is one of the popular methods for generating random samples. Popular methods under random design are simple random sampling \citep{gentle2003random,fang2005design}, Monte Carlo method (Markov Chain Monte Carlo sampling, Gibbs sampling, Metropolis-Hastings sampling, etc.) \citep{liu2001monte,robert2004monte,kroese2013handbook}. The main drawback of this method is that complete knowledge of the design space is needed. Also, this method shows poor coverage region or clustering for high dimension problems. Fig.~\ref{fig:random} illustrates the Gibbs sampling and the Metropolis-Hasting sampling for two factors. 

\subsection{Quasi-Random Design}

The quasi-random design has been widely used as a DoE sampling strategy. The samples generated using this scheme are neither fully random nor regular (like classical DoE schemes). Advantages of a quasi-random design are it uses the flexibility of random sampling techniques as well as the advantages of a grid scheme (as used in classical DoE designs). The samples using quasi-random design are uniform for high-dimensional problems, and they are also statistically dependent. Various techniques are available for quasi-random design, e.g. Latinized Centroidal Voronoi Tesselation sampling \citep{du1999centroidal,romero2006comparison}, \textit{maximin} Latin Hypercube sampling (LHS) \citep{stein1987large,chen2013optimizing,viana2016tutorial}, Sobol sampling \citep{bratley1988algorithm,owen1998scrambling,sobol2011construction}, Halton sampling \citep{chi2005optimal,weerasinghe2016particle}, Hammersley sampling \citep{hokayem2003quasi,dai2009application}, Faure sampling \citep{faure1982discrepance}, are commonly used quasi-random design, which are discussed briefly in the next section.

\subsubsection{Centroidal Voronoi Tesselation Sampling}

Centroidal Voronoi tesselation (CVT) is commonly used as a space-filling DoE sampling technique in applied mathematics, engineering and computer science field \citep{du1999centroidal}. In this technique, the generators, i.e. a set of points and a distance function, are considered. A Voronoi tesselation is a subset of the points which are closer to one of the generators than other generators \citep{ju2002probabilistic}. CVT is the tesselation when the generators coincide with the center of the mass of each subset. In general, the generators become the centroids of the Voronoi cell in which the region of the cell is defined by the subset. The algorithm of generating sampling using CVT begins with selecting location of the generators chosen randomly. The actual location of the generators are estimated iteratively and summarized in Algorithm~\ref{algo:CVT}. The convergence criteria of this iteration is usually taken as the difference of the centroids obtained in two successive iteration. The samples generated using CVT is shown in Fig.~\ref{fig:LCVT}.

 \begin{algorithm*}[h!]
 \caption{Pseudo-code for generation of CVT Sampling \citep{ju2002probabilistic}} \label{algo:CVT}
 \begin{algorithmic}[1]
 \State Aim: Generate $n$ samples over the $N$-dimensional space which are located at centroids of the centroidal Voronoi tesselation.
 \State Define the number of iteration \big($N_{Iter}$\big) and the number of random points generated in every iteration ($m$).
 \State Set the parameters $\alpha_1$, $\alpha_2$, $\beta_1$ and $\beta_2$ which define how the centroids are updated in every iteration. Also, $\alpha_2$, $\beta_2$ $>$ 0, $\alpha_1 + \alpha_2 = 1$ and $\beta_1 + \beta_2 = 1$.
 \State Choose an initial set of random samples, $x_i$, $i = 1, 2, \ldots, n$ which are assumed centroids over the domain.
 \State Set $p_i$ = 1, $i = 1, 2, \ldots, n$, which represents the number of times every centroid is updated.
 \State \textbf{Begin} 
 \For {Iteration = 1 to $N_{Iter}$}
 \State Generate the random samples $y_k$, $k = 1, 2, \ldots, m$ according to pre-defined probability density function, $\rho(\mathbf{x})$
 \For {$i = 1$ to $n$}
 \State Collect the set $W_i$ of all sample $y_k$ which are closest to $x_i$ and estimate their mean $u_i$.
 \If{ $W_i$ = $\phi$ (i.e. null set)}
  \State Do nothing
  \Else
    \State Update the centroid, $x_i$ as,
    \begin{equation*}
        x_i = \frac{(\alpha_1 p_i + \beta_1) x_i + (\alpha_2 p_i + \beta_2) u_i }{p_i + 1}, \hspace{5mm} p_i = p_i + 1
    \end{equation*}
  \EndIf
 \EndFor
 \State If the convergence criteria is satisfied i.e., the difference of centroids obtained in two successive iteration is less than tolerance, terminate.
 \EndFor
 \end{algorithmic}
 \end{algorithm*}
 

\subsubsection{Maximin Latin Hypercube Sampling}

LHS is a popular DoE sampling strategy \citep{viana2016tutorial} for building surrogate models. To generate $n$ samples using LHS, all $N$ dimensions are divided into $n$ equally spaced intervals, and a coordinate value is randomly selected from each interval along a dimension, yielding $n$ coordinate values in all $N$ dimensions. Later, a sample point consisting of $N$ coordinate values is built by randomly selecting one coordinate from each dimension. Thus, a two-dimensional region called the Latin square is formed for every sample. Details of the conventional LHS strategy are shown in Algorithm~\ref{algo:B_LHS}. The main drawback of LHS is that there is no guarantee that the samples generated cover the entire design space uniformly. For this reason, \textit{maximin} criteria \citep{johnson1990minimax}, which minimum distance between two samples is maximized, is used during LHS sample generation. Algorithm~\ref{algo:LHS} illustrates the pseudo-code for the \textit{maximin} LHS strategy. The samples generated using \textit{maximin} criteria are shown in Fig.~\ref{fig:Maxmin_LHS}.

\begin{algorithm}[h!]
 \caption{Pseudo-code for generation of basic Latin Hypercube Sampling} \label{algo:B_LHS}
 \begin{algorithmic}[1]
 \State Aim: Generate $n$ samples over the $N$-dimensional space.
 \State Let $x_{ij}$ is the $j$-th coordinate of the $i$-th sample, $x_i$.
 \State \textbf{Begin} 
 \For {$j$ = 1 to $N$}
 \State Generate $P_j$, a random permutation of the set \{1, $\ldots$, n\}.
 \EndFor
 \For {$j = 1$ to $N$}
 \For {$i = 1$ to $n$}
 \State Generate $x_{ij} = \frac{P_j (i) - U_j (i)}{n}$, where $U_j$ is the uniform random number in between [0,1].
 \EndFor
 \EndFor
 \end{algorithmic}
 \end{algorithm}
 
 \begin{algorithm*}[h!]
 \caption{Pseudo-code for generation of Maximin Latin Hypercube Sampling \citep{kamath2021intelligent}} \label{algo:LHS}
 \begin{algorithmic}[1]
 \State Aim: Generate $n$ samples over the $N$-dimensional space in which the minimum distance between samples are maximized.
 \State Define the number of iteration \big($N_{Iter}$\big) and the number of interchanges for every iteration ($m$).
 \State \textbf{Begin} 
 \For {Iteration = 1 to $N_{Iter}$}
 \State Generate $n$ samples using basic Latin hypercube sampling, shown in Algorithm~\ref{algo:B_LHS}.
 \State Assume $p_1$ and $p_2$ are the indices of samples with minimum distance.
 \For {$i = 1$ to $m$}
 \State Select index random either $p_1$ or $p_2$.
 \State For a random integer, $r_{col}$ in [1, $\ldots$, $N$], and a random row, $r_{row}$ in [1, $\ldots$, $n$], interchange the values in column $r_{col}$, rows index and $r_{row}$.
 \State If the minimum distance of samples is increased, accept the interchanges and update the values of $p_1$ and $p_2$.
 \EndFor
 \State If the minimum distance between samples is smaller than previous iteration, terminate.
 \EndFor
 \end{algorithmic}
 \end{algorithm*}

\subsubsection{Sobol Sampling}

The Sobol sequence is commonly used quasi-random design, proposed by Sobol \cite{sobol1967distribution}. The samples are generated using Sobol sequence are drawn from a special binary fraction ($v_i^j$) of length $w$ bits where $i$ = 1, 2, $\ldots$, $w$ and $j$ = 1, 2, $\ldots$, $N$. $N$ is the dimension of the problem. The numbers $v_i^j$ are known as direction numbers. To generate direction numbers of dimension $j$, a primitive polynomial over the field $F_2$ with elements \{0,1\} is considered, which is expressed as

\begin{equation}
    p_j (x) = x^q + b_1 x^{q-1} + \cdots + b_{q-1} x + 1
\end{equation}
The direction numbers, $v_i^j$ in the dimension $j$ are generated using the following recurrence relation, is given by
\begin{equation}
\begin{split}
    v_i^j = b_1 v_{i-1}^j \oplus b_2 & v_{i-2}^j \oplus \cdots \oplus b_{q-1} v_{i-(q-1)}^j \oplus \\
    & v_{i-q}^j \oplus \big(v_{i-q}^j / 2^q\big) ;\hspace{3mm} i > q
\end{split}
\end{equation}
In the above equation, the notation $\oplus$ represents the bitwise XOR. Finally, in the dimension $j$, the Sobol sequence is written as
\begin{equation}
    x_n^j = a_1 v_1^j \oplus a_2 v_2^j \oplus \cdots \oplus a_w v_w^j
\end{equation}
where $n = \sum_{i=0}^{w} a_i 2^i$. The coefficient $a_i$ is the random number in between \{0,1\}. Fig.~\ref{fig:sobol} illustrates the samples generated using the Sobol sequence.

\subsubsection{Halton Sampling}

The Halton sampling, proposed by Halton \cite{halton1960efficiency}, is quasi-random designs. Consider, first $N$ prime numbers as $m_1$, $m_2$, $\ldots$, $m_N$. The Halton sequence for $N$-dimensional space can be expressed as

\begin{equation}
    x_n = \Big(\Phi_{m_1} (n), \ldots, \Phi_{m_i} (n), \ldots, \Phi_{m_N} (n)\Big)
\end{equation}
where $\Phi_{m_i} (n)$ is the $i$-th radical inverse function, which is expressed in the mathematical form as
\begin{equation}
    \Phi_{m_i} (n) = \sum_{j=0}^{l(i)} a_j (i,n) m_i^{-(j+1)}
\end{equation}
In the above equation, $a_j$ is the integer coefficient, and it is given as $a_j (i,n) \in [0, m_i -1]$. Also, the integer terms $n$ and $l$ are given by
\begin{equation}
    n = \sum_{j=0}^{l(i)} a_j (i,n) m_i^{j}, \hspace{9mm} l(i) = \ceil{\log _{m_i} n}
\end{equation}
The samples generated using the Halton sequence are only bounded in between [0,1$]^N$ which are spread more evenly within the design space compared to random sampling. Fig.~\ref{fig:halton} shows the samples generated using the Halton sequence. It is noted that the samples in Fig.~\ref{fig:halton} are scaled considering the bound [-1,1].


\subsubsection{Hammersley Sampling}

The Hammersley sampling is a popular quasi-random design, similar to the Halton sequence. The Hammersley sequence differs from the Halton sequence in that the Hammersley sequence employs ($N$-1), $\Phi$ sequence. The Hammersley sequence for $N$-dimensional space is expressed as
\begin{equation}
    x_n = \Big( \frac{n}{N}, \Phi_{m_1} (n), \ldots, \Phi_{m_i} (n), \ldots, \Phi_{m_{N-1}} (n)\Big)
\end{equation}

\begin{figure*}[h!]
\centering     
\subfigure[Centroidal Voronoi Tesselation] {\label{fig:LCVT}\includegraphics[width=0.49\textwidth]{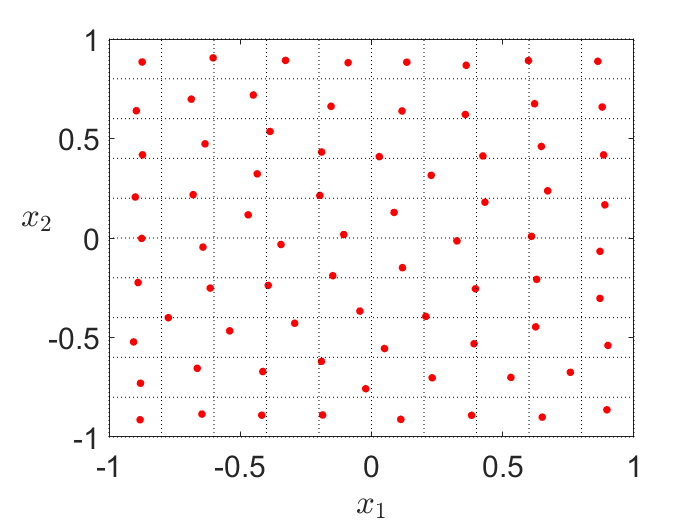}} 
\subfigure[Maximin Latin Hypercube]{\label{fig:Maxmin_LHS}\includegraphics[width=0.49\textwidth]{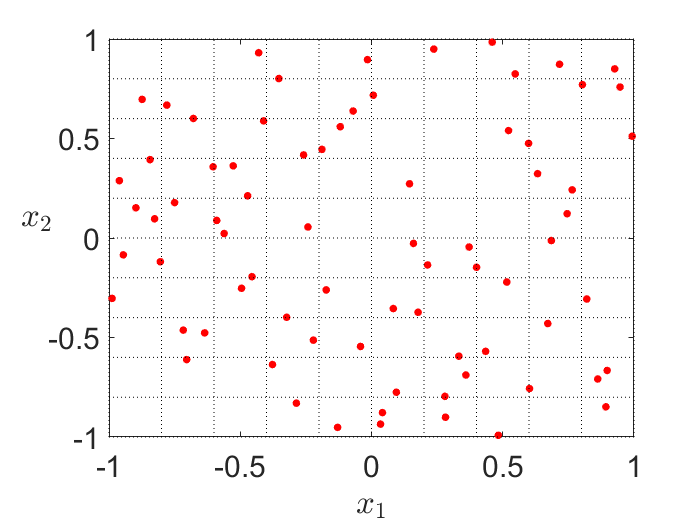}}
\subfigure[Sobol Samples]{\label{fig:sobol}\includegraphics[width=0.49\textwidth]{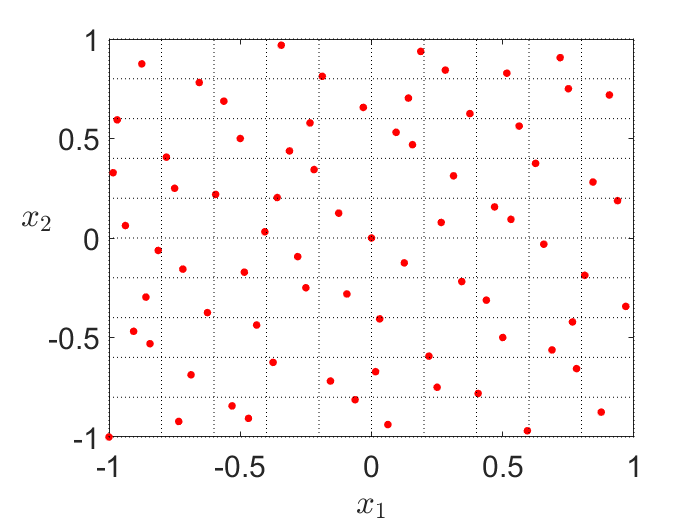}}
\subfigure[Halton Samples]{\label{fig:halton}\includegraphics[width=0.49\textwidth]{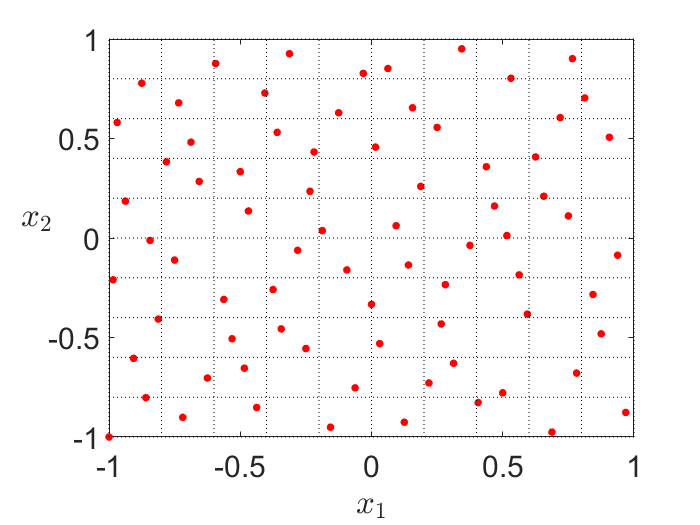}}
\subfigure[Hammersley Samples]{\label{fig:hammersley}\includegraphics[width=0.49\textwidth]{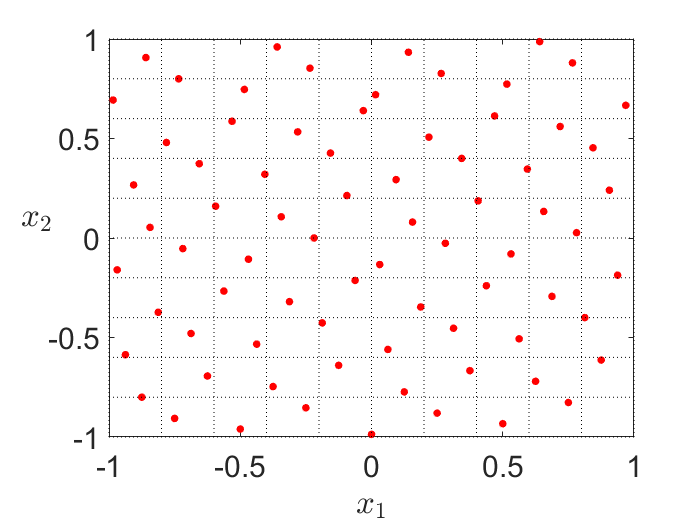}}
\subfigure[Faure Samples]{\label{fig:faure}\includegraphics[width=0.49\textwidth]{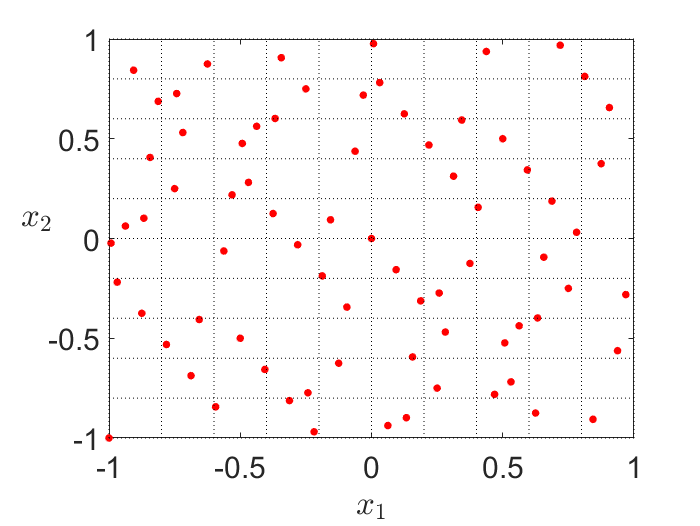}}
\caption{Various sampling strategies for quasi-random design}
\label{fig:quasi}
\end{figure*}

The primary distinction between the Halton and Hammersley sequences is that the Halton sequence can generate an infinite number of samples (i.e., infinite $N$), whereas the Hammersley sequence requires an upper bound on the number of samples. Fig.~\ref{fig:hammersley} shows the samples generated using the Hammersley sequence.


\subsubsection{Faure Sampling}

The Faure sequence, which is similar to the Halton sequence, proposed by Faure \cite{faure1982discrepance}, is a well-known quasi-random sample generation technique. Consider, $m$ is the first prime number such that $m \geq n$, where $n$ is the total number of samples. Also, the upper bound of the sample size is $m^p$. This is the main difference between the Faure and the Halton sequences. For example, if $n$ = 50 i.e., 50 samples are required to generate, the last Halton sequence (i.e., in dimension 50) uses the 50-th prime number, which is 229, whereas the Faure sequence uses the first prime number after 50. The main advantage of using Faure sampling over Halton sampling is that the Faure sequence is faster to fill the gaps for high dimensional problems. Also, it prevents correlation problems in high dimensions, which are seen in the Halton sequence. 

Let $c_{ij}$ = $\binom{i}{j}$ (mod $m$) where $0 \leq j \leq i \leq p$. This implies $\big\{c_{ij} - \binom{i}{j}\big\}$ is the multiple of $m$. The base $m$ representation of $n$ is expressed as
\begin{equation}
    n = \sum_{i=0}^{p-1} a_i (n) m^i
\end{equation}
where the coefficients $a_i \in [0, m)$ takes the integer values. The samples produced by the Faure sequence are denoted by
\begin{equation}
    x_n = \sum_{j=0}^{p-1} a_j (n) m^{-(j+1)}
\end{equation}
where the coefficient $a_j = \sum_{l=j}^{p-1} c_{lj} a_l(n)$ (mod $p$), $j \in \{0, 1, \ldots, p-1\}$. Fig.~\ref{fig:faure} shows the samples generated using the Faure sequence.


\subsection{Full Grid Design}

Full grid design, also known as the collocation technique, is a well established scheme for generating sample points \citep{tatang1996direct,mathelin2005stochastic,xiu2005high}. In this study, a Lagrangian polynomial is used for the interpolation. Let, $x_i^j$ be the support nodes which are expressed as $x_i^j \in x^j = \{x_1^j, x_2^j, \ldots, x_{n_j}^j\} \in$ [-1, 1] be a sequence of abscissas for Lagrange interpolation on [-1, 1]. The collocation method is based on one-dimensional Lagrange polynomial interpolation at $N$ Gauss quadrature points and is expressed as
\begin{equation}\label{eq:collocation1}
    \mathcal{U}^j (y) = \sum_{i=1}^{n_j} y\Big(x_i^j\Big) \cdot \mathcal{L}_i^j (x)
\end{equation}
where $n_j$ is the total number of grid points. Also, in the above equation, $\mathcal{L}$ is the Lagrange polynomial, given by
\begin{equation}
    \mathcal{L}_i^j (x) = \prod_{k=1,k \neq i}^{n_j} \Bigg(\frac{x - x_k^j}{x_i^j - x_k^j}\Bigg)
\end{equation}
Eq.~\ref{eq:collocation1} defines the one-dimensional problem. For multi-variate cases, the full tensor product interpolation is written as
\begin{equation}\label{eq:collocation2}
\begin{split}
    \mathcal{I}^\beta \mathcal{U} (y) & = \Big(\mathcal{U}^{j_1} \otimes \mathcal{U}^{j_2} \otimes \cdots \otimes \mathcal{U}^{j_\beta} \Big)(y) \\
    &= \sum_{i_1=1}^{n_1} \sum_{i_2=1}^{n_2} \cdots \sum_{i_\beta=1}^{n_\beta} y \Big(\mathbf{x}_{i_1}^{j_1} , \mathbf{x}_{i_2}^{j_2}, \ldots, \mathbf{x}_{i_\beta}^{j_\beta} \Big) \cdot \\
    &\hspace{10mm}\Big(\mathcal{L}_{i_1}^{j_1}(\mathbf{x}) \otimes \mathcal{L}_{i_2}^{j_2}(\mathbf{x}) \otimes \cdots \otimes \mathcal{L}_{i_\beta}^{j_\beta}(\mathbf{x}) \Big)
\end{split}
\end{equation}
\begin{figure*}[h!]
\centering     
\subfigure[$\mu$ = 1, $\beta$ = 2] {\label{fig:col1}\includegraphics[width=0.32\textwidth]{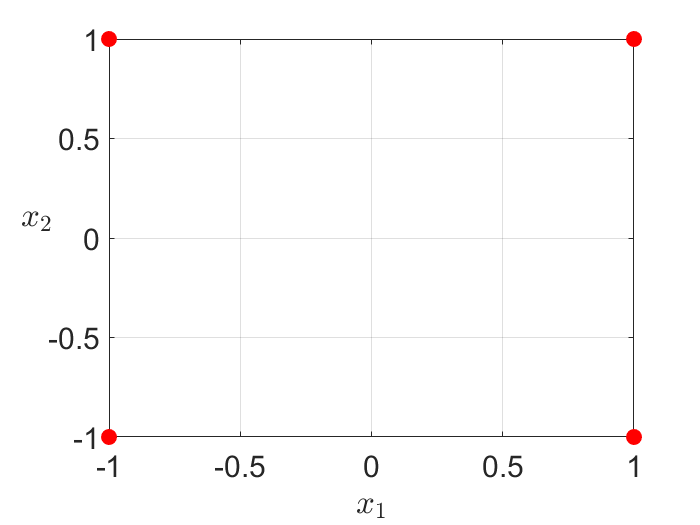}}
\subfigure[$\mu$ = 2, $\beta$ = 2] {\label{fig:col2}\includegraphics[width=0.32\textwidth]{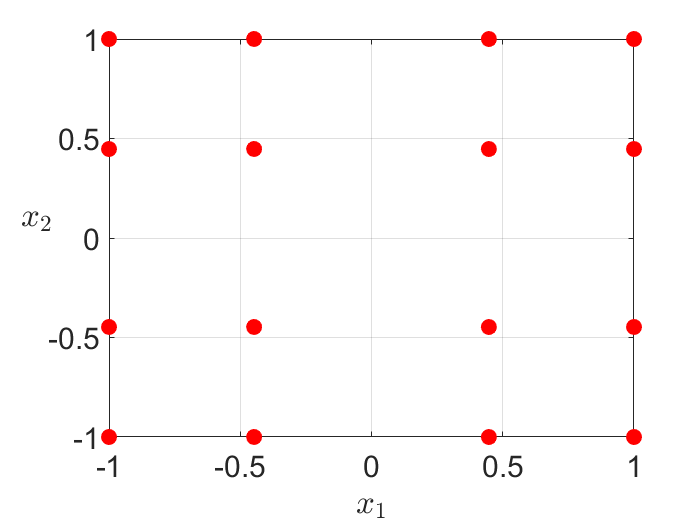}}
\subfigure[$\mu$ = 3, $\beta$ = 2]
{\label{fig:col3}\includegraphics[width=0.32\textwidth]{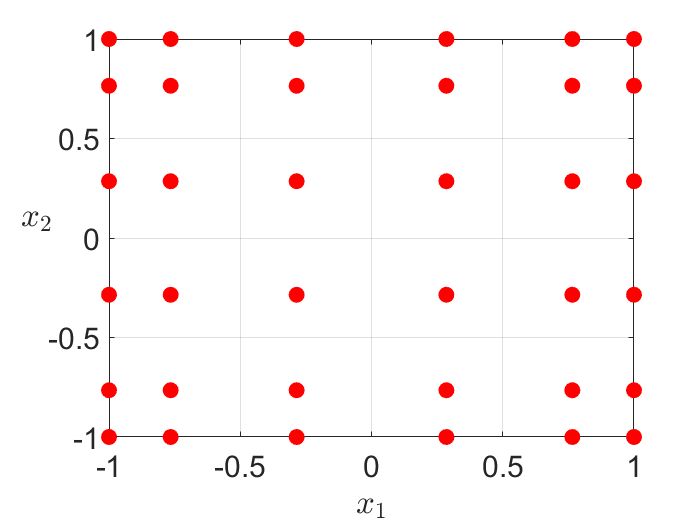}}
\subfigure[$\mu$ = 1, $\beta$ = 3] {\label{fig:col1_3d}\includegraphics[width=0.32\textwidth]{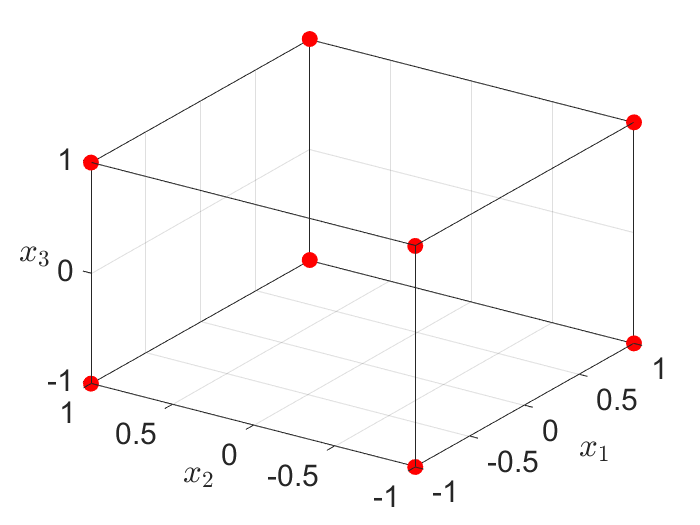}}
\subfigure[$\mu$ = 2, $\beta$ = 3] {\label{fig:col2_3d}\includegraphics[width=0.32\textwidth]{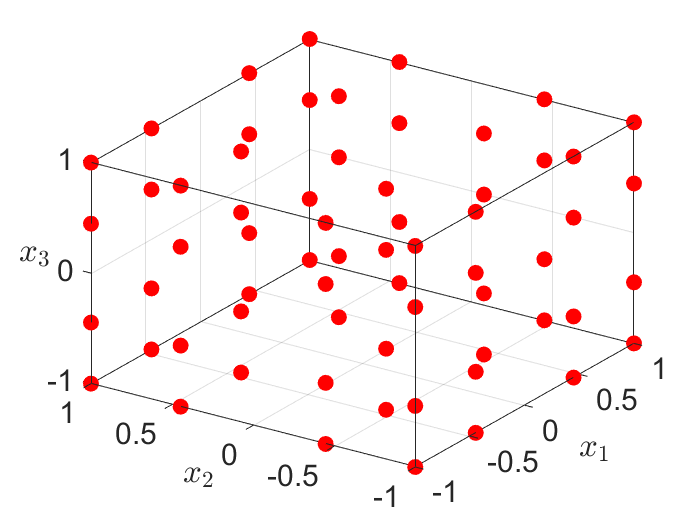}}
\subfigure[$\mu$ = 3, $\beta$ = 3] {\label{fig:col3_3d}\includegraphics[width=0.32\textwidth]{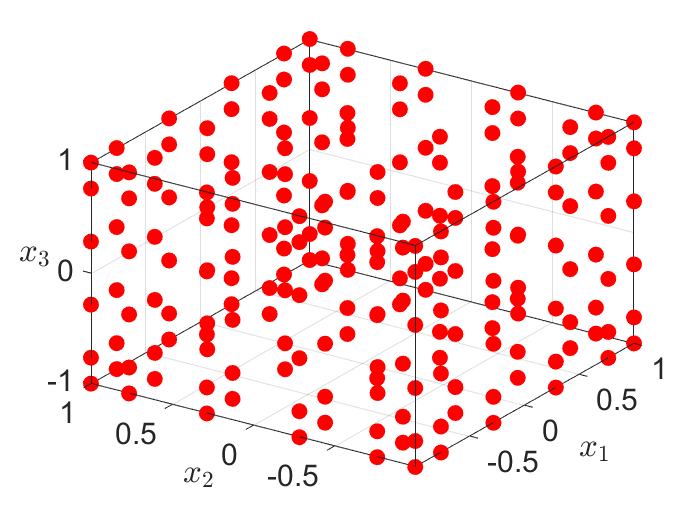}}
\caption{Full tensor product grid for different levels for two and three factors}
\label{fig:collocation}
\end{figure*}
where $\beta$ is the dimension of the problem. Fig.~\ref{fig:collocation} illustrates the grid points generated using the collocation technique for various levels ($\mu$) of sparse grid for two and three random variables.


\subsection{Sparse Grid Design}\label{sec:sgd}

\begin{figure*}[h!]
\centering     
\subfigure[$\mu$ = 1, $\beta$ = 2] {\label{fig:2d_lev1}\includegraphics[width=0.32\textwidth]{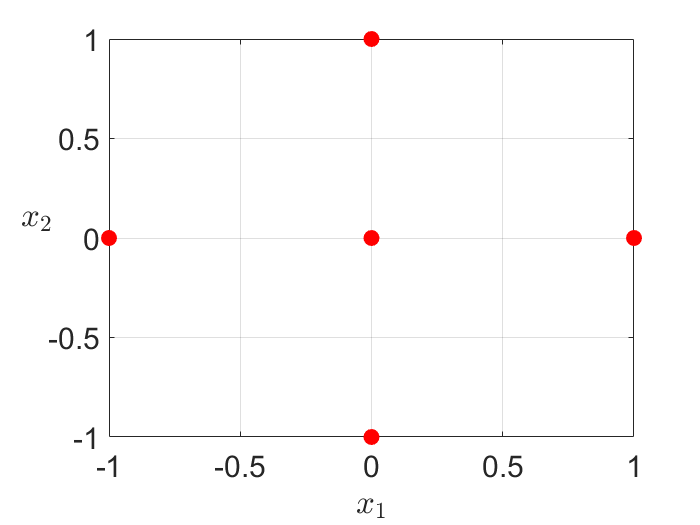}}
\subfigure[$\mu$ = 2, $\beta$ = 2] {\label{fig:2d_lev2}\includegraphics[width=0.32\textwidth]{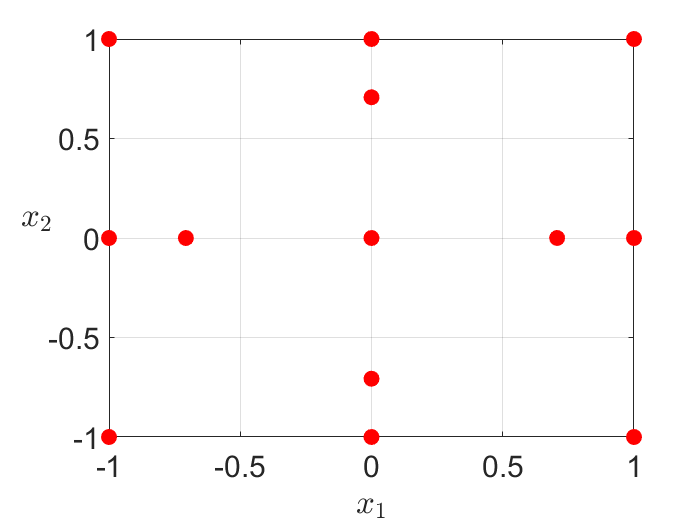}}
\subfigure[$\mu$ = 3, $\beta$ = 2] {\label{fig:2d_lev3}\includegraphics[width=0.32\textwidth]{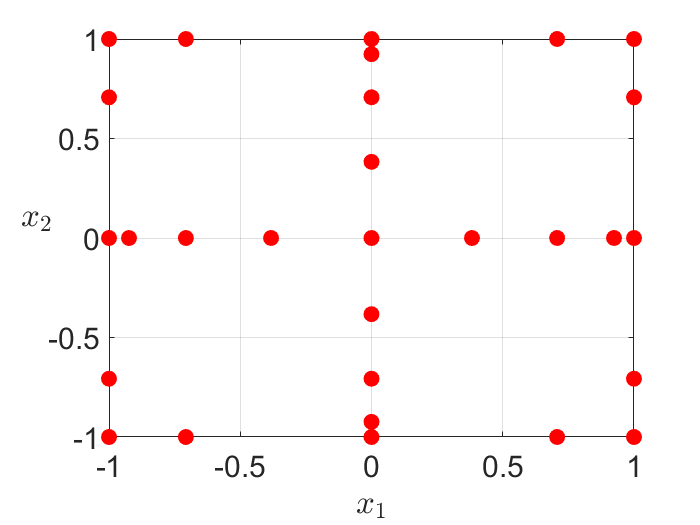}}
\subfigure[$\mu$ = 1, $\beta$ = 3] {\label{fig:3d_lev1}\includegraphics[width=0.32\textwidth]{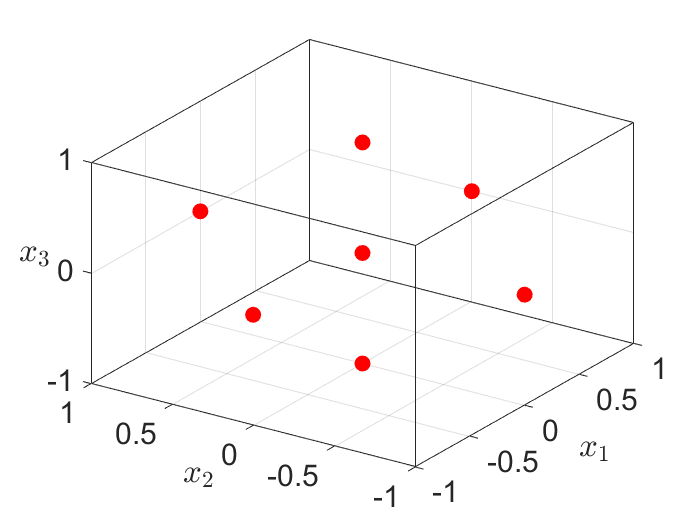}}
\subfigure[$\mu$ = 2, $\beta$ = 3] {\label{fig:3d_lev2}\includegraphics[width=0.32\textwidth]{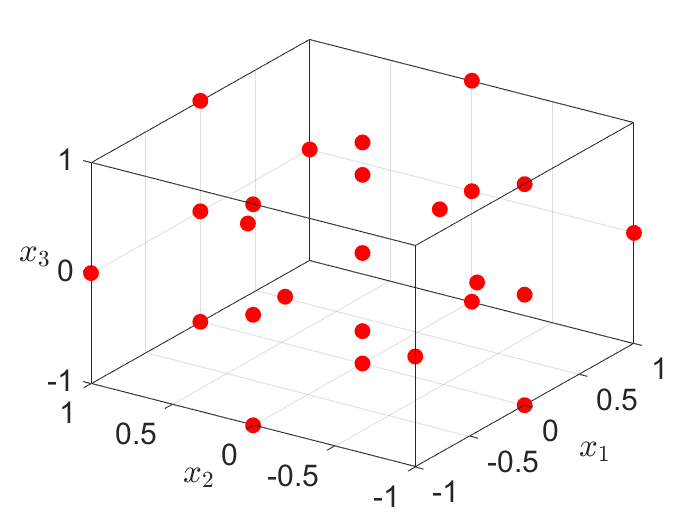}}
\subfigure[$\mu$ = 3, $\beta$ = 3] {\label{fig:3d_lev3}\includegraphics[width=0.32\textwidth]{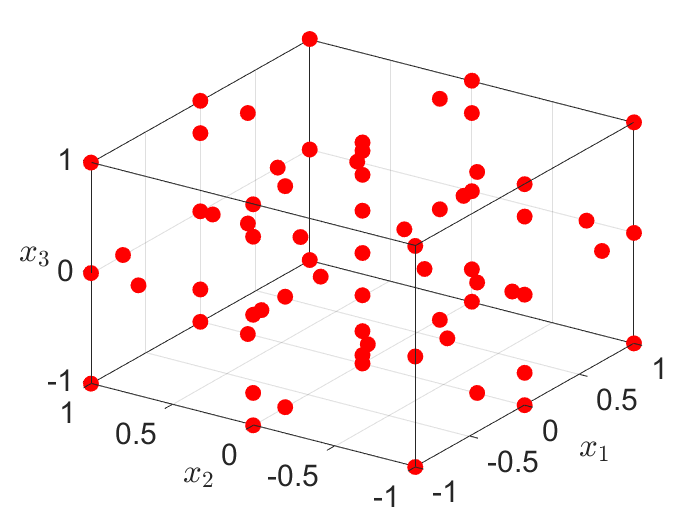}}
\caption{Grid points generated using sparse grid scheme}
\label{fig:sparse_grid}
\end{figure*}

The sparse grid scheme has been proven an efficient scheme for generating support points \citep{holtz2010sparse}. In this study, Smolyak's algorithm \citep{klimke2005computing} is utilized to generate sampling, based on the hierarchical sampling scheme \citep{bungartz2004sparse}. The samplings generated from previous interpolation depth are used in the subsequent interpolation to produce the new sampling set. The sparse grid ($\mathcal{SG}$) can be formulated as
\begin{equation}\label{eq:sp1}
    \mathcal{SG}_{\mu} = \sum_{\sum_{i=1}^\beta j_i \leq \mu+\beta-1} \Big(\Delta^{j_1} \otimes \Delta^{j_2} \otimes \cdots \otimes \Delta^{j_\beta} \Big) \big(y\big)
\end{equation}
where $\mu$ is the level of the sparse grid design and $\beta$ denotes the dimension of the problem, i.e., the number of variables considered. In Eq.~\ref{eq:sp1}, the difference function, $\Delta^j$ equals to ($\mathcal{U}^j - \mathcal{U}^{j-1}$) where $\mathcal{U}$ is the interpolation function for the univariate case. The unidimensional interpolation function can be expressed using quadrature form, is given by
\begin{equation}\label{eq:sp2}
    \mathcal{U}^j (y) = \sum_{i=1}^{n_j} y\Big(x_i^j\Big) \cdot \Breve{A}_i^j (x)
\end{equation}
In the above equation, $n_j$ denotes the number of grid points for a given random parameter. Also, $\Breve{A}_i^j$ is the basis function, which is $\Breve{A}_i^j \in \mathcal{C} ([-1,1])$, $\Breve{A}_i^j(x_k^j) = \delta_{ik}$. Here, $x_i^j$ are the support nodes, which are expressed as $x_i^j \in x^j = \{x_1^j, x_2^j, \ldots, x_{n_j}^j\}$ and $x_k^j \in [-1, 1]$, $1 \leq k \leq n_j$. The interpolation function for the univariate case, defined in Eq.~\ref{eq:sp2}, can be extended to the multivariate case as:
\begin{equation}\label{eq:sp3}
\begin{split}
   & \Big(\mathcal{U}^{j_1} \otimes \mathcal{U}^{j_2} \otimes \cdots \otimes \mathcal{U}^{j_\beta} \Big)(y) \\
   & = \sum_{i_1=1}^{n_1} \sum_{i_2=1}^{n_2} \cdots \sum_{i_\beta=1}^{n_\beta} y \Big(\mathbf{x}_{i_1}^{j_1} , \mathbf{x}_{i_2}^{j_2}, \ldots, \mathbf{x}_{i_\beta}^{j_\beta} \Big) \cdot\\
   & \hspace{14mm} \Big(\Breve{A}_{i_1}^{j_1}(\mathbf{x}) \otimes \Breve{A}_{i_2}^{j_2}(\mathbf{x}) \otimes \cdots \otimes \Breve{A}_{i_\beta}^{j_\beta}(\mathbf{x}) \Big)
\end{split}
\end{equation}
The sparse grid interpolation presented above uses two types of basis functions, i.e., piecewise linear and polynomial basis functions. In this study, the Chebyshev-Gauss-Lobatto grid \citep{barthelmann2000high} is used, which is one of the most commonly used polynomial grids (i.e., higher order basis functions are considered) and provides the higher accuracy. In the Chebyshev-Gauss-Lobatto grid scheme, the total number of grid points for each random variable and grid points are given as:
\begin{subequations}
\begin{equation}
    n_j = \begin{cases}
    1 , \hspace{13mm}\text{if} \hspace{2mm} j = 1 \\
    2^{j-1}+1 ,\hspace{2mm} \text{if} \hspace{2mm} j > 1 
    \end{cases}
\end{equation}
\begin{equation}
    x_i^j = - \cos \frac{\pi(i-1)}{n_j -1}, \hspace{5mm} i = 1, 2, \ldots, n_j
\end{equation}
\end{subequations}
Also, in Eq.~\ref{eq:sp3}, the basis function, $\Breve{A}_i^j$ is given by
\begin{equation}
    \Breve{A}_i^j  = \begin{cases}
    \!\begin{aligned}
    \frac{2}{n_j -1}& \Big[ 1 - \frac{\cos (\pi(i-1))}{n_j(n_j-2)} -\\&  2 \sum_{k=1}^{(n_j-3)/2} \frac{1}{4k^2-1} 
     \cdot \cos \frac{2\pi k (i-1)}{n_j -1}\Big], \\
     &\hspace{15mm}\text{for} \hspace{3mm} i = 2, \ldots, (n_j-1)\end{aligned} \\
    \frac{1}{n_j(n_j-2)}, \hspace{11mm} \text{for} \hspace{3mm} i = 1 \& n_j
    \end{cases}
\end{equation}
Fig.~\ref{fig:sparse_grid} shows the grid points generated using a sparse grid scheme for various levels ($\mu$) of sparse grid for two and three random variables. 

\subsection{Rotational Sparse Grid Design}

\begin{figure}[h!]
\centering     
\subfigure[]{\label{fig:2d_lev3_m}\includegraphics[width=0.49\textwidth]{figures/2d_lev3.png}}
\subfigure[]{\label{fig:2d_lev3_rot}\includegraphics[width=0.49\textwidth]{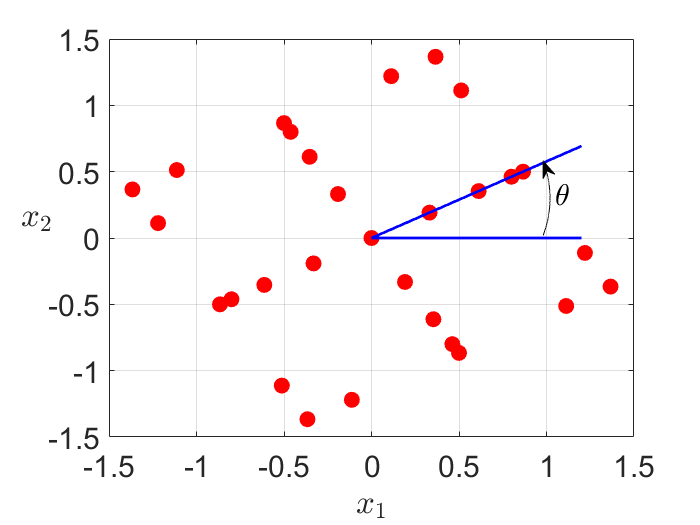}}
\caption{Support points generated using \protect\subref{fig:2d_lev3_m} sparse grid and \protect\subref{fig:2d_lev3_rot} rotational sparse grid scheme for two dimensional problem}. 
\label{fig:Rot_sparse_grid}
\end{figure}

Recently, rotational sparse grid design has been developed as one of the efficient design of experiment schemes, proposed by Wu \textit{et al.} \cite{wu2021reliability}. In this method, the initial support points are generated through a sparse grid scheme, as mentioned in Section~\ref{sec:sgd}. To obtain maximum information from those support points, an optimization has been performed by rotating all support points with a fixed angle of rotation. The rotational matrix is given by for the two-dimensional problem as:
\begin{equation}
    \mathcal{R} (\theta) = \begin{bmatrix}
    \cos{\theta} & - \sin{\theta}\\
    \sin \theta &  \cos \theta
    \end{bmatrix}
\end{equation}
where $\theta$ is the angle of rotation. Similarly, for three-dimensional problems, the rotational matrix is written as:
\begin{subequations}
\begin{equation}
    \mathcal{R}_x (\theta) = \begin{bmatrix}
    1 & 0 & 0\\
    0 & \cos{\theta} & - \sin{\theta}\\
    0 & \sin \theta &  \cos \theta
    \end{bmatrix}
\end{equation}
\begin{equation}
    \mathcal{R}_y (\theta) = \begin{bmatrix}
     \cos{\theta} & 0 &  \sin{\theta}\\
     0 & 1 & 0\\
     -\sin{\theta} & 0 &  \cos{\theta}
     \end{bmatrix}
\end{equation}
\begin{equation}
     \mathcal{R}_z (\theta) = \begin{bmatrix}
     \cos{\theta} & - \sin{\theta} & 0\\
     \sin{\theta}  &  \cos{\theta} & 0\\
     0  & 0 & 1\\
    \end{bmatrix}
\end{equation}
\end{subequations}
In the above equation, $\mathcal{R}_x (\theta)$ is the rotational matrix where the rotation is performed with respect to the x-axis. Similarly, $\mathcal{R}_y (\theta)$ and $\mathcal{R}_z (\theta)$ are the rotational matrix for rotations about the y- and z-axes, respectively. For \textit{n}-dimensional problem, the rotational matrix is given by
\begin{equation}
   \begin{matrix} \mathcal{R} (\theta) =  \begin{bmatrix}
    1 & \cdots &  0 & \cdots  & 0 & \cdots  & 0\\
    \vdots & \ddots & \vdots & \ddots & \vdots & \ddots & \vdots\\
    0 & \cdots & \cos{\theta}& \cdots & - \sin{\theta} & \cdots  & 0\\
    \vdots & \ddots & \vdots & \ddots & \vdots & \ddots & \vdots\\
    0 & \cdots & \sin \theta & \cdots &  \cos \theta & \cdots  & 0 \\
    \vdots & \ddots & \vdots & \ddots & \vdots & \ddots & \vdots\\
    0 & \cdots &  0 & \cdots  & 0 & \cdots  & 0
    \end{bmatrix} \\
\text{\textit{i}-th} \hspace{5mm} \text{\textit{j}-th} \hspace{9mm}  \text{\textit{k}-th}
    \end{matrix} 
    \begin{array}{c}
 \text{\textit{i}-th} \vspace{6mm} \\ \vspace{8mm} \text{\textit{j}-th} \\ \vspace{8mm} \text{\textit{k}-th}\\ \\ \\
\end{array}
\end{equation}
The support points on $j$-th and $k$-th axes are rotated with respect to the $i$-th axis in this case. The grid points using a sparse grid scheme are shown in Fig.~\ref{fig:Rot_sparse_grid} for a two-dimensional problem. The angle of rotation ($\theta$) is estimated by performing the optimization problem considering some objective functions from the designer's perspective.

\section{Machine Learning} \label{sec:ML}
\begin{figure*}[h!]
    \centering
    \includegraphics[width=1\textwidth]{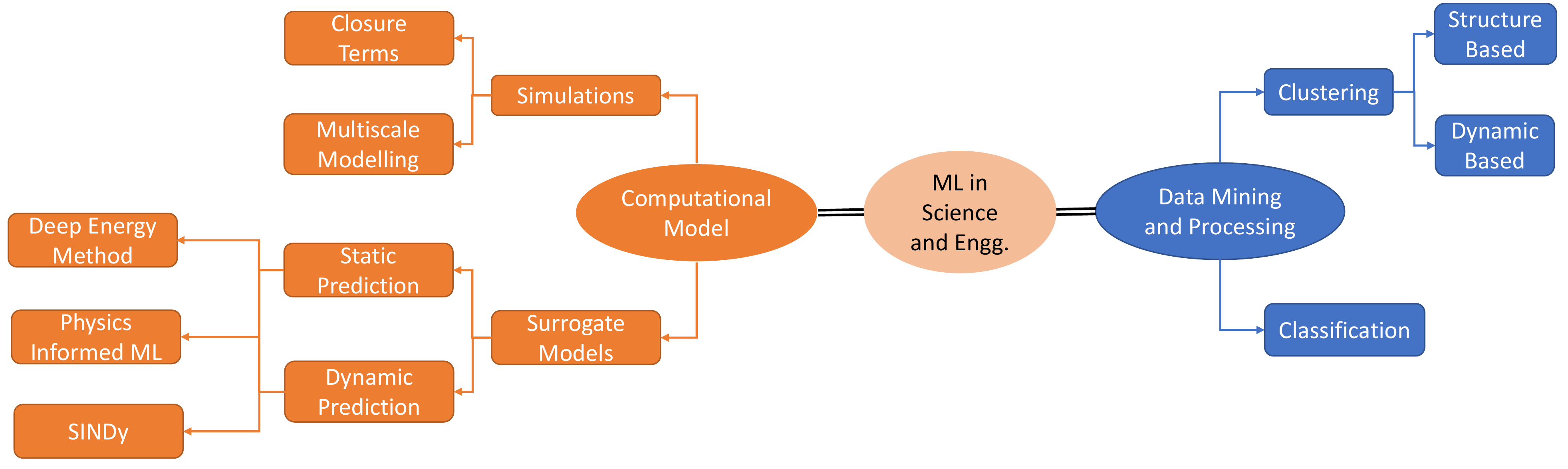}
    \caption{An overview of applications of machine learning}
    \label{fig:ML_app}
\end{figure*}

Machine learning is categorized into four groups, i.e., supervised learning, unsupervised learning, semi-supervised learning, and reinforcement learning. Supervised learning is used to develop the predicted models where there is a mapping between the set of input vectors and one or more outputs. Regression and classification problems belong to supervised learning. Unsupervised learning is used to estimate the probability distribution of the provided data. Prediction using this algorithm is not possible because the data is not labeled. Clustering is an example of unsupervised learning. Semi-supervised learning is the combination of supervised and unsupervised learning where few samples are labelled (i.e., supervised) and unsupervised learning is used to enhance the performance of supervised learning. Finally, the reinforced learning interacts with the environment to learn the behaviour of the decision variable by enhancing the expected average reward \citep{burkov2019hundred}. Fig.~\ref{fig:ML_app} illustrates the applications of machine learning in the science and engineering fields.

\subsection{Feed-Forward Neural Networks}\label{sec:FNN}

\begin{figure*}[h!]
    \centering
    \includegraphics[width=0.7\textwidth]{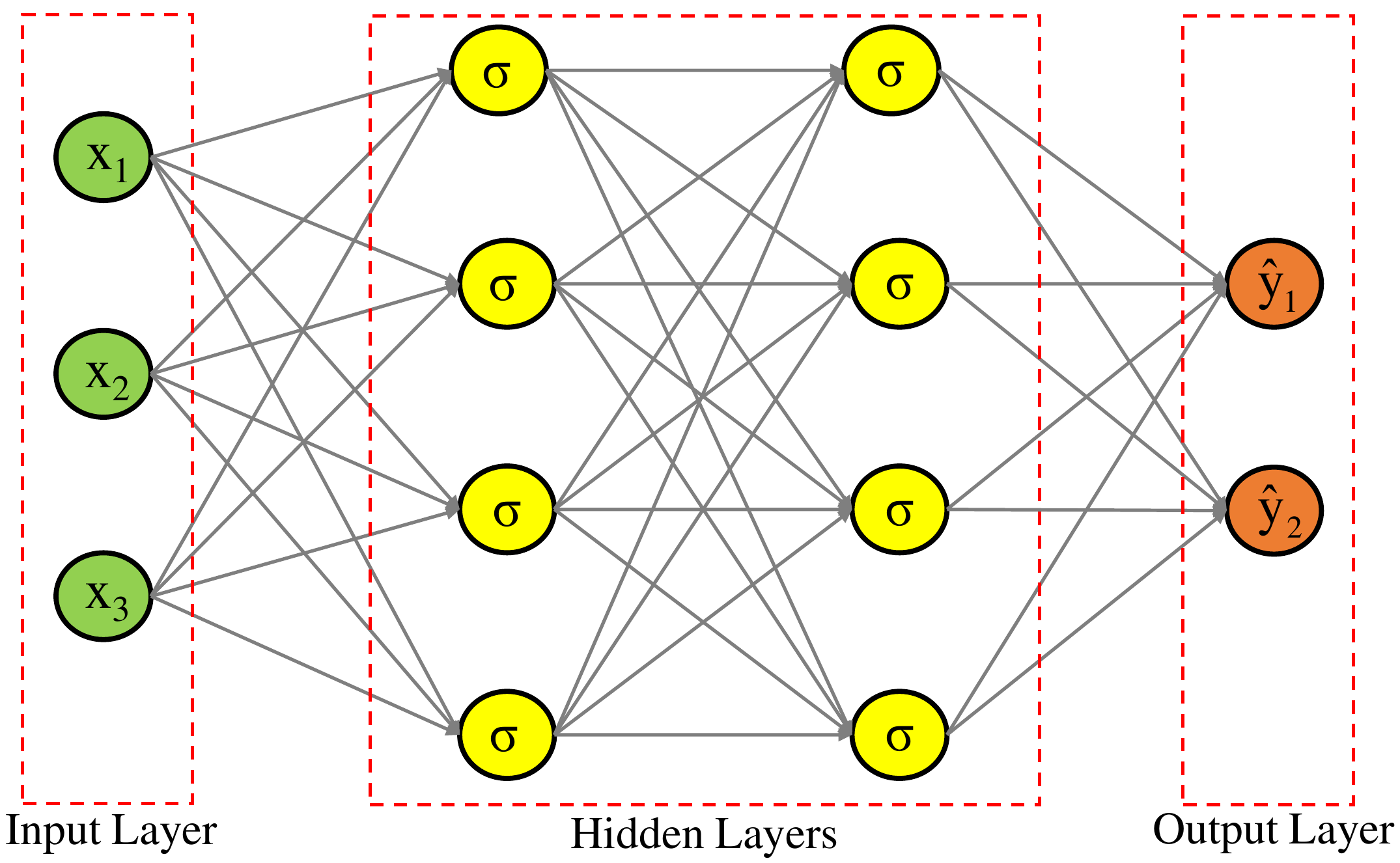}
    \caption{Basic architecture of feed-forward neural network}
    \label{fig:ann}
\end{figure*}

Artificial neural networks belong to the supervised learning class, which mimics the neurons of the brain. Generally, it is a mapping between input and output vectors. It consists of at least three layers, i.e., the input layer, the hidden layer, and the output layer. As the information propagates from the input layer through hidden layers to the output layer, it is known as feed-forward neural networks. When more than one hidden layer is used, the network is called ``deep'' neural network. In each hidden layer, one or more neurons are used. 
A schematic diagram of the architecture of the neural network is shown in Fig.~\ref{fig:ann}. 
The output of the network is a linear combination of the neurons, which is related to the weight of the connection, called synaptic weight, and a transfer function, called the activation function.

From Fig.~\ref{fig:ann}, $\mathbf{x}$ = [$x_1$, $x_2$, $\ldots$, $x_n$] are the $n$ input variables. The output of a neuron is estimated to be two steps. Firstly, the input vector $\mathbf{x}$ is transformed into a linear combination, i.e.,
\begin{equation}
    z = \mathbf{w}^T \mathbf{x} + b
\end{equation}
where $\mathbf{w}$ is the synaptic weight and $b$ is the bias. Second, the neuron output $a(\mathbf{x})$ is estimated by feeding $z$ into an activation function, $\sigma$, in order to obtain the data's nonlinearity, which is given as
\begin{equation}
    a(\mathbf{x}) = \sigma (z)
\end{equation}
The common activation functions used in neural networks are tabulated in Table~\ref{tab:active_fun}. For accurate prediction of the output response ($y$), each synaptic weight in each layer needs to be updated. The backpropagation learning algorithm is used for updating the weights. In that case, the loss function ($\mathcal{L}$) is taken into account and must be minimized. The commonly used loss functions are tabulated in Table~\ref{tab:loss_fun}. The stochastic gradient descent technique is used to solve the minimization problem, which is expressed as
\begin{equation}
    [\mathbf{w}]^o = [\mathbf{w}]^{o-1} - \eta \frac{\partial \mathcal{L}}{\partial \mathbf{w}} 
\end{equation}
where $o$ is the current iteration and $\eta$ represents the learning rate.

\begin{table*}[h!]
\small
    \centering
    \caption{Common activation functions used in neural networks}
    \begin{tabular}{c c c c}
    \hline
       No.  & Activation Function & Equation  &  Range \\
       \hline
      1  & Aranda \citep{gomes2013optimization} & $\sigma (x) = 1 - (1+2e^x)^{-1/2}$ & [0 , 1]  \\
      2  &  Bi-sig1 \citep{sodhi2014bi} & $\sigma (x) = 0.5\Big(\frac{1}{1+e^{-x+1}} + \frac{1}{1+e^{-x-1}}\Big)$ &  [0 , 1] \\
      3  &  Bi-sig2 \citep{sodhi2014bi} & $\sigma (x) = 0.5\Big(\frac{1}{1+e^{-x}} + \frac{1}{1+e^{-x-1}}\Big)$   & [0 , 1]  \\
      4  & Bi-tanh1 \citep{sodhi2014bi} & $\sigma (x) = 0.5\Big[\tanh(\frac{x}{2}) + \tanh(\frac{x+1}{2}) \Big] + 0.5$ &  [-0.5 , 1.5] \\
      5  & Bi-tanh2 \citep{sodhi2014bi} & $\sigma (x) = 0.5\Big[\tanh(\frac{x-1}{2}) + \tanh(\frac{x+1}{2}) \Big] + 0.5$ &  [-0.5 , 1.5]   \\
      6  &  Cloglog \citep{gomes2008complementary} &  $\sigma (x) = 1 - e^{-e^x}$ &  [0 , 1]    \\
      7  & Cloglogm \citep{gomes2011comparison} &  $\sigma (x) = 1 - 2e^{-0.7e^x} + 0.5$ & [-0.5 , 1.5]   \\
      8  &  Elliott \citep{elliott1993better} & $\sigma (x) = \frac{0.5x}{1+|x|} +0.5$ & [0 , 1]    \\
      9  & ELU & $\sigma (x) = \begin{cases} \alpha (e^x-1),& x < 0\\
      x,& x \geq 0
      \end{cases}$ & [-$\infty$ , +$\infty$]    \\
      10  &  Gaussian & $\sigma (x) = e^{-x^2}$ & [0 , 1]   \\
      11  &  Heaviside & $\sigma (x) = \begin{cases} 0,& x < 0\\
      0.5,& x = 0\\
      1,& x > 0
      \end{cases}$ & [-$\infty$ , +$\infty$]  \\
      12  &  Hyperbolic Tangent & $\sigma (x) = \frac{e^x -e^{-x}}{e^x +e^{-x}}$ & [-$\infty$ , +$\infty$]  \\
      13  &  Linear & $\sigma (x) = x$ & [-$\infty$ , +$\infty$]  \\
      14  & Logarithmic & $\sigma (x) = \begin{cases} \ln(1+x)+0.5,& x \geq 0\\
      -\ln (1-x)+0.5,& x < 0
      \end{cases}$ & [-$\infty$ , +$\infty$]  \\
      15  & Loglog \citep{gomes2011comparison} &  $\sigma (x) = e^{-e^{-x}} + 0.5$ &  [0.5 , 1.5]   \\
      16  & Logsigm \citep{singh2003class+} & $\sigma (x) = \Big(\frac{1}{1+e^{-x}}\Big)^2 +0.5$ &  [0.5 , 1.5] \\
      17  & Log-sigmoid & $\sigma (x) = \frac{1}{1+e^{-x}}$ &  [0 , 1] \\
      18  & Modified Elliott \citep{burhani2015denoising} & $\sigma (x) = \frac{x}{\sqrt{1+x^2}} + 0.5$ & [-0.5 , 1.5]   \\
      19  &  Piece-wise Linear & $\sigma (x) = \begin{cases} 1,& x \geq 0.5\\
      x+0.5,& -0.5< x< 0.5\\
      0,& x \leq -0.5
      \end{cases}$ & [-$\infty$ , +$\infty$]  \\
      20  & PReLU & $\sigma (x) = \begin{cases} \alpha x,& x < 0\\
      x,& x \geq 0
      \end{cases}$ & [-$\infty$ , +$\infty$]    \\
      21  & ReLU & $\sigma (x) = \begin{cases} 0,& x < 0\\
      x,& x \geq 0
      \end{cases}$ & [-$\infty$ , +$\infty$]    \\
      22  & Rootsig \citep{duch1999survey} & $\sigma (x) = \frac{x}{1+\sqrt{1+x^2}} + 0.5$ & [-0.5 , 1.5]   \\
      23  & Saturated  & $\sigma (x) = \frac{|x+1| - |x-1|}{2} + 0.5$ & [-0.5 , 1.5]  \\
      24  & Sech & $\sigma (x) = \frac{2}{e^x + e^{-x}}$ & [0 , 1]  \\
      25  &  Sign & $\sigma (x) = \begin{cases} -1,& x < 0\\
      0,& x = 0\\
      1,& x > 0
      \end{cases}$ & [-$\infty$ , +$\infty$]  \\
      26  & Sigmoidalm \citep{singh2003class+} & $\sigma (x) = \Big(\frac{1}{1+e^{-x}}\Big)^4 + 0.5$ &  [0.5 , 1.5] \\
      27  &  Sigmoidalm2 \citep{chandra2004case} & $\sigma (x) = \Big(\frac{1}{1+e^{-x/2}}\Big)^4 + 0.5$ &  [0.5 , 1.5]   \\
      28  & Sigt \citep{yuan2013new} & $\sigma (x) = \frac{1}{1+e^{-x}} + \frac{1}{1+e^{-x}} \Big(1 -\frac{1}{1+e^{-x}}\Big)$ &[0 , 1]   \\
      29  & Skewed-sig \citep{chandra2014skewed} & $\sigma (x) = \Big(\frac{1}{1+e^{-x}}\Big) \Big(\frac{1}{1+e^{-2x}}\Big) + 0.5$ &  [0.5 , 1.5] \\
      30  & Softplus & $\sigma (x) = \ln (1+e^x)$ & [-$\infty$ , +$\infty$]    \\
      31  &  Softsign \citep{elliott1993better} & $\sigma (x) = \frac{x}{1+|x|} + 0.5$ & [-0.5 , 1.5]  \\
      32  &  Wave \citep{hara1994comparison} & $\sigma (x) =(1-x^2)e^{-x^2}$  &  [-0.055 , 1] \\
      \hline
    \end{tabular}
    \label{tab:active_fun}
\end{table*}

\begin{table*}[h!]
\centering
\begin{threeparttable}
    \caption{Common loss functions used in neural networks for regression analysis}
    \label{tab:loss_fun}
    \begin{tabular}{c c c }
    \hline
       No.  & Loss Function & Equation   \\
       \hline
       1 & Mean absolute loss & $||y-\hat{y}||_1$ \\
       2 & Mean squared loss & $||y-\hat{y}||^2_2$ \\
       3 & Expectation loss & $||y-$ {p}$(\hat{y})||_1$ \\
       4 & Regularised expectation loss & $||y-$ p$(\hat{y})||^2_2$ \\
       5 & Chebyshev loss & $\mathrm{max_j} \mid$ p $(\hat{y})^{(j)} - y^{(j)}\mid$\\
       6 & Hinge loss & $\sum_{j} \mathrm{max} \Big(0, \frac{1}{2} - y^{(j)} \hat{y}^{(j)} \Big)$ \\
        7 & Squared hinge loss & $\sum_{j} \mathrm{max} \Big(0, \frac{1}{2} - y^{(j)} \hat{y}^{(j)} \Big)^2$ \\
        8 & Cubed hinge loss & $\sum_{j} \mathrm{max} \Big(0, \frac{1}{2} - y^{(j)} \hat{y}^{(j)} \Big)^3$ \\
        9 & Cross entropy loss & $- \sum_{j} y^{(j)} \log \mathrm{p}(\hat{y})^{(j)} $ \\
        10 & Squared log loss & $- \sum_{j} \Big[y^{(j)} \log \mathrm{p}(\hat{y})^{(j)}\Big]^2 $ \\
        11 & Tanimoto loss & -$\frac{\sum_{j} \mathrm{p}(\hat{y})^{(j)} y^{(j)}} {||\mathrm{p}(\hat{y})||_2^2 + ||y||_2^2 - \sum_{j} \mathrm{p}(\hat{y})^{(j)} y^{(j)}}$\\
        12 & Cauchy-Schwarz Divergence & $- \log \frac{\sum_{j} \mathrm{p}(\hat{y})^{(j)} y^{(j)}}{||\mathrm{p}(\hat{y})||_2 ||y||_2}$\\
         \hline
    \end{tabular}
    \begin{tablenotes}
      \small
      \item $y$ and $\hat{y}$ represent true and predicted value
      \item $j$ represents the dimension of a given vector
      \item $\mathrm{p}(\cdot)$ is the probability estimate
    \end{tablenotes}
  \end{threeparttable}
\end{table*}

\subsection{Physics-Informed Neural Network}

\begin{figure*}[h!]
    \centering
    \includegraphics[width=1\textwidth]{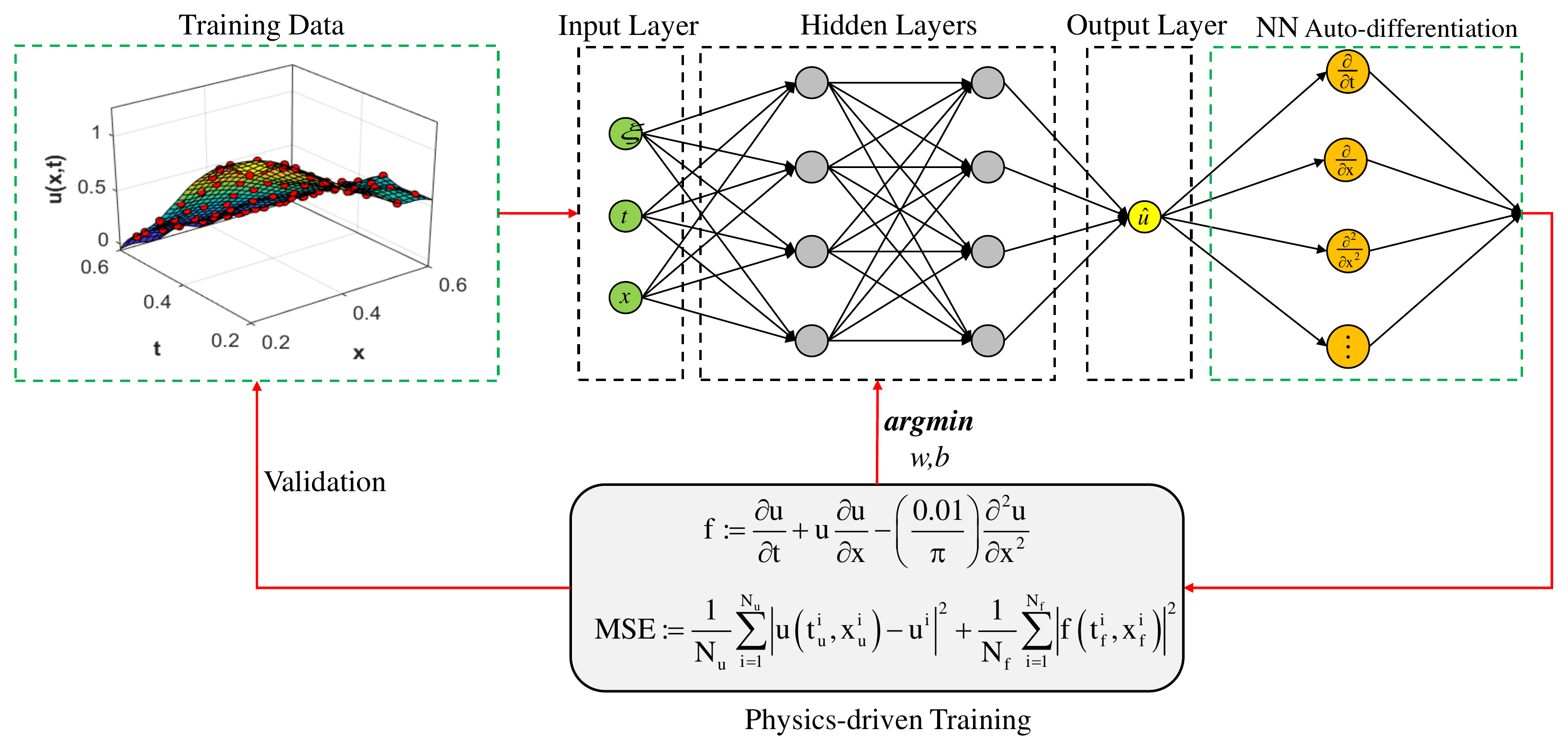}
    \caption{Schematic diagram of physics informed neural network}
    \label{fig:pinn1}
\end{figure*}

A physics-informed neural network (PINN) is a special class of feed-forward neural network where physical conditions are imposed on the neural network as a loss function. The data-driven neural network, as discussed in Section~\ref{sec:FNN}, is often cost-effective for a complex physical system where a large amount of solution data is required. PINN is used to solve problems where only a few data points are available, such as noisy data from an experiment. To estimate the accurate solution of the physical system, the physical laws of the governing system are imposed on the network. To illustrate the algorithm of PINN, a nonlinear partial differential equation is considered, which is expressed in the following form
\begin{equation}\label{eq:pinn1}
    \frac{\partial u}{\partial t} + \mathcal{N} [u; \lambda] = 0; \hspace{4mm} x \in \Omega , t \in \mathcal{T}
\end{equation}
The latent solution in the above equation is $u(x,t)$, which is a function of time $t \in [0 , T]$ and a spatial variable $x \in \Omega$, where $\Omega$ represents a space in $\mathbb{R}^D$. Also, in Eq.~\ref{eq:pinn1}, $\mathcal{N} [u; \lambda]$ denotes a non-linear differential operator with coefficients $\lambda$. Like in feed-forward neural networks, the weights and bias in PINN are estimated by minimizing a cost function, which is given as
\begin{equation}
    \mathcal{L} = {\mathcal{L}}_u + {\mathcal{L}}_f
\end{equation}
where ${\mathcal{L}}_u$ and ${\mathcal{L}}_f$ denote the residual of a PDE (i.e., mean square error between $u(x,t)$ and $\hat{u}(x,t)$) and the residual corresponding to initial and boundary conditions, respectively. 
PINN is used for either data-driven interference or data-driven identification of  differential equations. Data-driven interference is a forward problem in which $\lambda$ of the differential equation (Eq.~\ref{eq:pinn1}) is known and $u(x,t)$ is solved based on initial and boundary conditions, whereas data-driven identification identifies coefficients $\lambda$ of the differential equation based on scattered data of the solution $u(x,t)$. A schematic diagram of PINN is shown in Fig.~\ref{fig:pinn1}.

\section{Numerical Results}\label{sec:numerical}

In this section, numerical examples are presented to illustrate the performance of a PINN considering various design experiment schemes to generate the training data-set. For this purpose, five different nonlinear differential equations are considered, i.e., Viscous Burger's equation, Shr\"{o}dinger equation, one-dimensional heat equation, Allen-Cahn equation and Korteweg-de Vries equation. The neural network is designed to solve those PDEs, which approximates the true solution $u(t,x)$, denoted by $\hat{u}(t,x)$. The number of neurons in the hidden layers is constant for a given problem. The total number of hidden layers and the number of neurons in each hidden layer are obtained by comparing the prediction responses using a neural network with the actual solution multiple times. The hyperbolic tangent tanh is used as an activation function for every neuron and this is same for all problems. To train the neural network, different DoE schemes are used, i.e., factorial design (FD), central composite design (CCD), centroidal Voronoi tesselation (CVT), maximin Latin hypercube sampling (MLH), Sobol sampling, Halton sampling, Hammersley sampling, Faure sampling, full grid design (FGD), and sparse grid design (SGD). The accuracy of the predicted responses using a neural network depends on the number of DoE samples in the training set. Therefore,  preliminary study is devoted to its accuracy by carrying out the simulation considering increasing size of the training set  (i.e., from 50 to 500 with a step size of 50). The accuracy is evaluated using the mean squared error (MSE) between actual and predicted responses. For the simulation purposes, \texttt{python} using \texttt{TensorFlow} \citep{abadi2016tensorflow} and \texttt{SciANN} \citep{haghighat2021sciann}, \texttt{MATLAB} softwares are used.


\subsection{Problem 1: Viscous Burger's Equation}

As the first example, viscous Burger's equation is illustrated, whose applications are seen in the fluid mechanics, acoustics, and traffic flow fields \citep{dafermos2005hyperbolic}. The governing PDE equation for viscous Burger's equation is given by \citep{raissi2018deep}
\begin{equation}\label{eq:burger1}
        \frac{\partial u}{\partial t} + u \frac{\partial u}{\partial x} - \nu \pdv[2]{u}{x} = 0
\end{equation}
where $(x,t) \in \Xi$ = [-1 , 1] $\times$ [0 , 1]; $\nu$ is the viscosity of the system, which is taken as (0.01/$\pi$). The initial and boundary conditions of the above equation are given by
\begin{equation}
    u (t = 0, x) = -\text{sin}(\pi x); \hspace{2mm} u (t, x = \pm 1) = 0 
\end{equation}
With the above equation, a function $f(t,x)$ is constructed which provides the physical information of the neural network. It is expressed as follows:
\begin{equation}\label{eq:bur11}
    f(t,x) = \frac{\partial u}{\partial t} + u \frac{\partial u}{\partial x} - \nu \pdv[2]{u}{x}
\end{equation}
The optimal parameters of the neural network are estimated by minimizing the loss function, which is given by
\begin{equation}
     {\mathcal{L}} =  {\mathcal{L}}_u +  {\mathcal{L}}_f
\end{equation}
where ${\mathcal{L}}_u$ denotes mean square error loss, which is calculated using the data corresponding to the initial and boundary conditions, whereas ${\mathcal{L}}_f$ denotes the loss corresponding to the function $f(t,x)$, defined in Eq.~\ref{eq:bur11}. These two loss functions are expressed in the following forms
\begin{subequations}\label{eq:bur12}
\begin{equation}
    {\mathcal{L}}_u = \frac{1}{N_u} \sum_{i=1}^{N_u} \Bigl\vert u (t_u^i,x_u^i) - u^i \Bigl\vert^2
\end{equation}
\begin{equation}
    {\mathcal{L}}_f = \frac{1}{N_f} \sum_{i=1}^{N_f} \Bigl\vert f (t_f^i,x_f^i) \Bigl\vert^2
\end{equation}
\end{subequations}
In Eq.~\ref{eq:bur12}, $\{t_u^i,x_u^i,u^i\}$ denotes the initial and boundary training data of the function $u(t,x)$ where $N_u$ is the total number of training data. Similarly, $\{t_f^i,x_f^i\}$ is the training data of the spatio-temporal domain where $N_f$ is the total number of training data to train $f(t,x)$.

\begin{figure}[h!]
    \centering
    \includegraphics[width=0.49\textwidth]{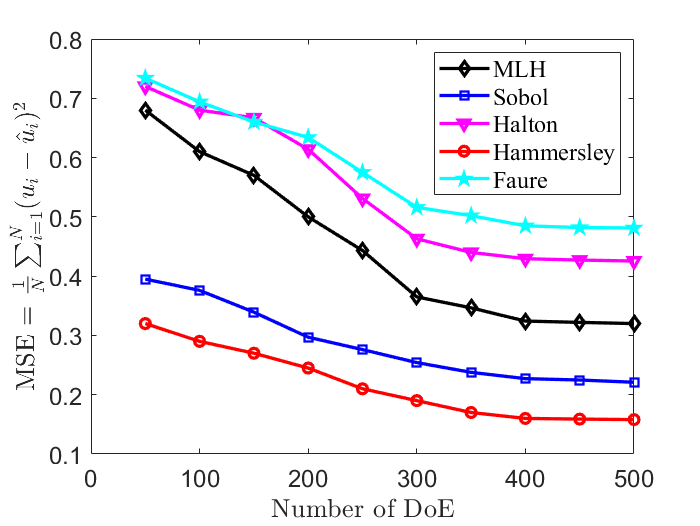}
    \caption{The MSE history for different levels of DoE for solving Viscous Burger's equation}
    \label{fig:conv_burger}
\end{figure}

In this problem, eight hidden layers, each with 20 neurons are used construct the neural network. The MSE values for various sizes of DoE sets are shown in Fig.~\ref{fig:conv_burger}. From Fig.~\ref{fig:conv_burger}, it is clear that 400 samples are needed to stabilize the MSE values.

\begin{figure*}[h!]
    \centering
    \includegraphics[width=1.0\textwidth]{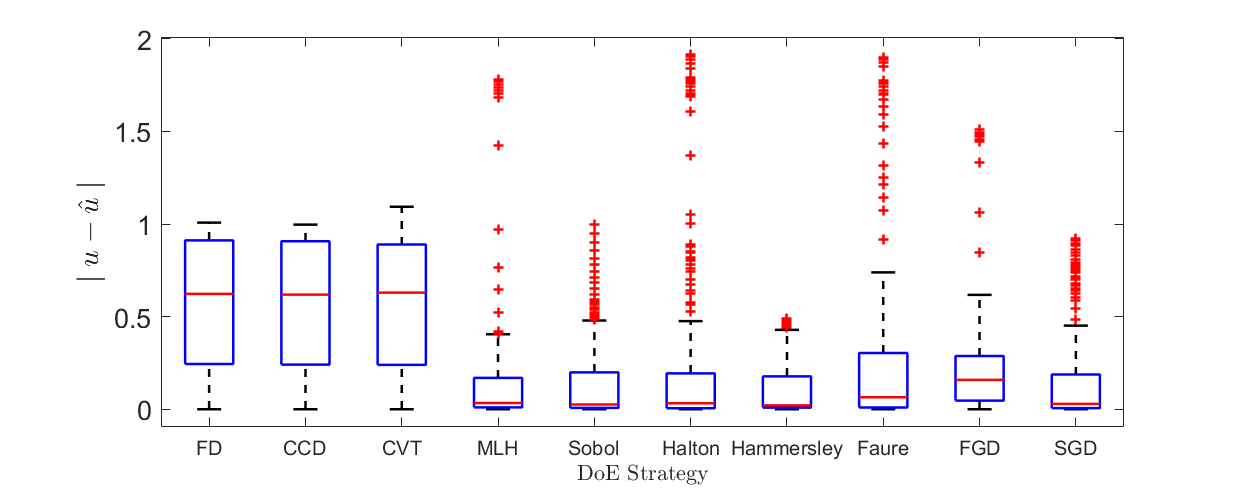}
    \caption{Absolute error between original and predicted responses of viscous Burger's equation considering different DoE strategies}
    \label{fig:burger_error}
\end{figure*}
\begin{figure*}[h!]
\centering     
\subfigure[]{\label{fig:exp1}\includegraphics[width=0.45\textwidth]{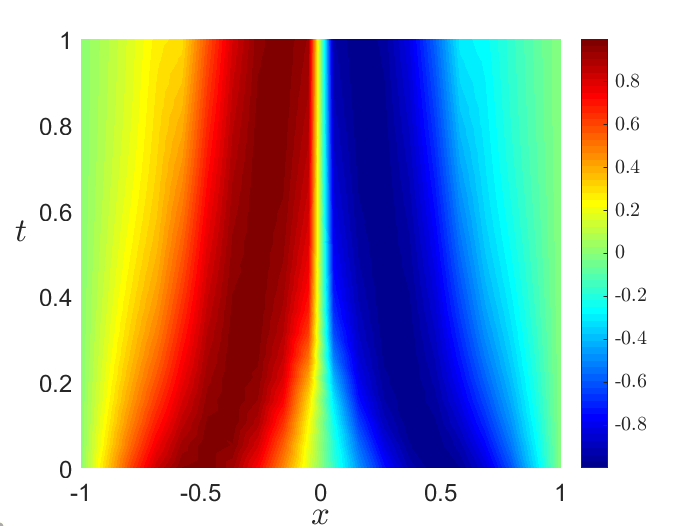}}
\subfigure[]{\label{fig:exp2}\includegraphics[width=0.45\textwidth]{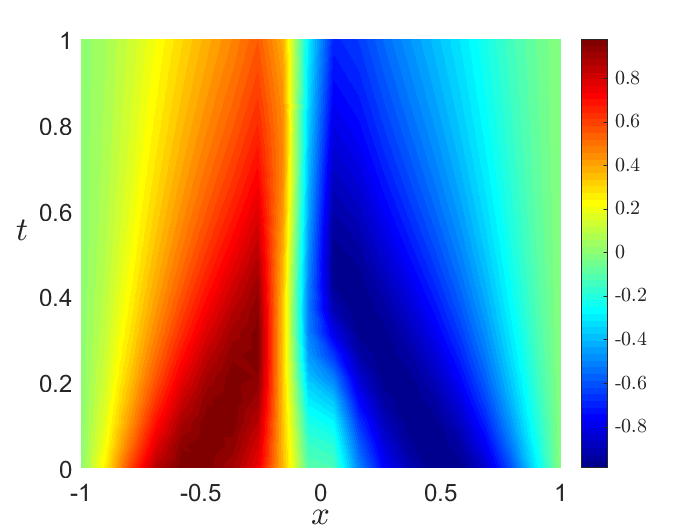}}
\subfigure[]{\label{fig:exp3}\includegraphics[width=0.45\textwidth]{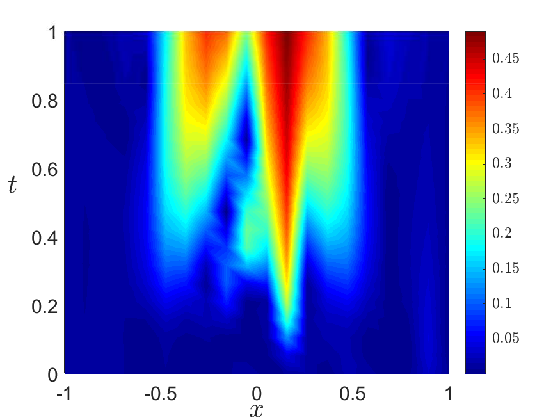}}
\caption{Viscous Burger's equation: \protect\subref{fig:exp1} Exact solution; \protect\subref{fig:exp2} Solution using PINN and \protect\subref{fig:exp3} Absolute error between exact solution and solution using PINN}
\label{fig:burger1}
\end{figure*}
With this, a comparative study is performed with all DoE schemes and the absolute errors between actual and predicted responses are shown using a Box and Whisker plot as shown in Fig.~\ref{fig:burger_error}. In Fig.~\ref{fig:burger_error}, the red line represents the median value of the errors. By comparing the median values, it can be concluded that FD, CCD, and CVT perform the worst over-fitting. Among those, Hammersley performs best due to a lower median value. SGD and Sobol sampling yield the next best results. Once the accuracy is established, the predicted responses are evaluated considering the 400 DoE samples generated using Hammersley sampling. The contour plots of the actual solution ($u(t,x)$) and predicted solution ($\hat{u}(t,x)$) of the viscous Burger's equation are shown in Fig.~\ref{fig:exp1} and Fig.~\ref{fig:exp2}. The error between these two solutions is shown in Fig.~\ref{fig:exp3}.


\subsection{Problem 2 : Shr\"{o}dinger Equation}

Here, the Shr\"{o}dinger equation is considered as our next example, which is a classical field equation, applied in quantum mechanical systems, nonlinear wave propagation in optical fibres, etc. The nonlinear Shr\"{o}dinger equation subjected to periodic boundary conditions is written as \citep{raissi2019physics}:
\begin{equation}\label{eq:scrodinger}
    \begin{split}
        & i \frac{\partial u}{\partial t} + \frac{1}{2} \pdv[2]{u}{x} + |u|^2 u  = 0 \\
        u (0,x) & = 2 \text{sech}(x) ; \hspace{3mm}
        u (t, -5)  = u (t,5) \\
        &\hspace{2mm} \frac{\partial u}{\partial x} (t, -5)  = \frac{\partial u}{\partial x} (t, 5)
    \end{split}
\end{equation}
where $(x,t) \in \Xi$ = [-5, 5] $\times$ [0, $\pi$/2]. In the above equation, $u (x,t)$ is a complex-valued function. A loss function is considered, which has to be minimized to estimate the parameters related to neural networks, is given by
\begin{equation}
    \mathcal{L} = {\mathcal{L}}_0 + {\mathcal{L}}_b + {\mathcal{L}}_f
\end{equation}
where ${\mathcal{L}}_0$, ${\mathcal{L}}_b$, and ${\mathcal{L}}_f$ are the residual loss functions on the initial data, the periodic boundary condition and Shr\"{o}dinger equation, are written as
\begin{subequations}
\begin{equation}
    {\mathcal{L}}_0 = \frac{1}{N_0} \sum_{j=1}^{N_0}  \Bigl\vert u(0, x_0^j) - u_0^j\Bigl\vert^2 
\end{equation}
\begin{equation}\label{eq:sch_loss2}
\begin{split}
    {\mathcal{L}}_b = \frac{1}{N_b}  \sum_{j=1}^{N_b} \bigg\{ &\Bigl\vert u^j (t_b^j, -5) - u^j (t_b^j, 5) \Bigl\vert^2 + \\
    &\Bigl\vert u_x^j (t_b^j, -5) - u_x^j (t_b^j, 5) \Bigl\vert^2 \bigg\}
\end{split}
\end{equation}
\begin{equation}\label{eq:sch_loss3}
    {\mathcal{L}}_f = \frac{1}{N_f} \sum_{j=1}^{N_f} \Bigl\vert f\big(t_f^j, x_f^j\big) \Bigl\vert^2 
\end{equation}
\end{subequations}
In Eq.~\ref{eq:sch_loss2}, $u_x$ denotes the first partial derivative of $u$ with respect to $x$, i.e., $\partial u / \partial x$. Also, in Eq.~\ref{eq:sch_loss3}, $f$ is the function representing Shr\"{o}dinger equation, as in Eq.~\ref{eq:scrodinger} i.e., $f = i u_t + 0.5 u_{xx}+ |u|^2u$. 

\begin{figure}[h!]
    \centering
    \includegraphics[width=0.49\textwidth]{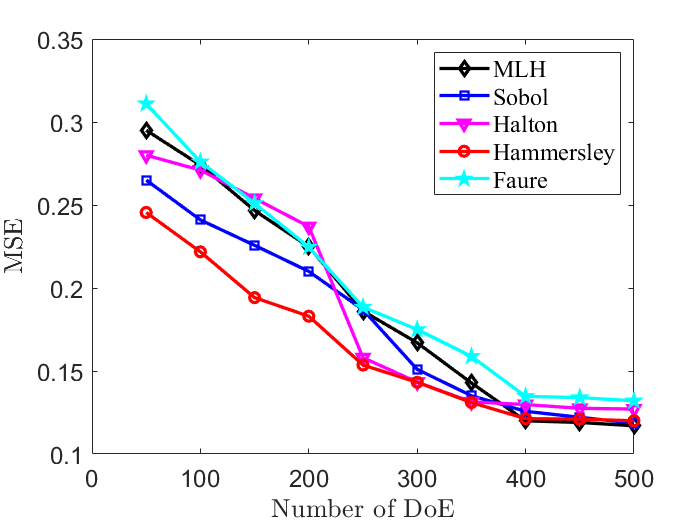}
    \caption{The MSE values for different number of DoE sets for different DoE techniques for Shr\"{o}dinger equation}
    \label{fig:conv_shrodinger}
\end{figure}
\begin{figure*}[h!]
\centering     
\subfigure[]{\label{fig:shr1}\includegraphics[width=0.49\textwidth]{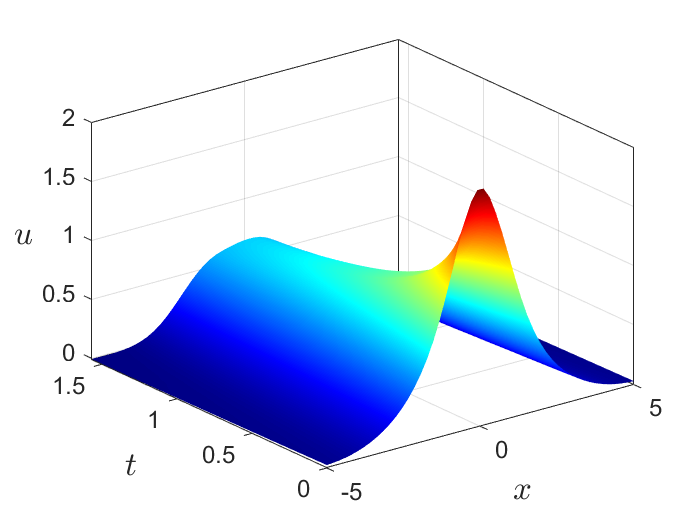}}
\subfigure[]{\label{fig:shr2}\includegraphics[width=0.49\textwidth]{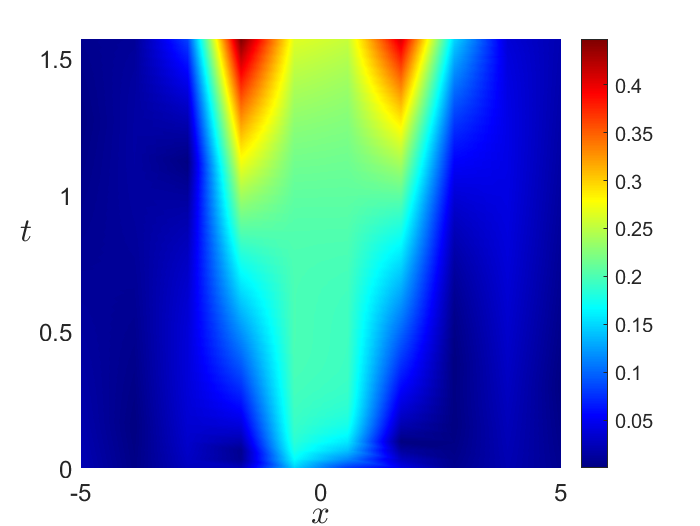}}
\caption{Shr\"{o}dinger equation: \protect\subref{fig:shr1} Solution for $u$ using PINN and \protect\subref{fig:shr2} Absolute error between exact solution and solution using PINN}
\label{fig:shrodinger1}
\end{figure*}

Fig.~\ref{fig:conv_shrodinger} depicts the MSE values for various DoE set sizes. It is seen that 400 samples are needed to stabilize the MSE values. 
\begin{figure*}[h!]
    \centering
    \includegraphics[width=1.0\textwidth]{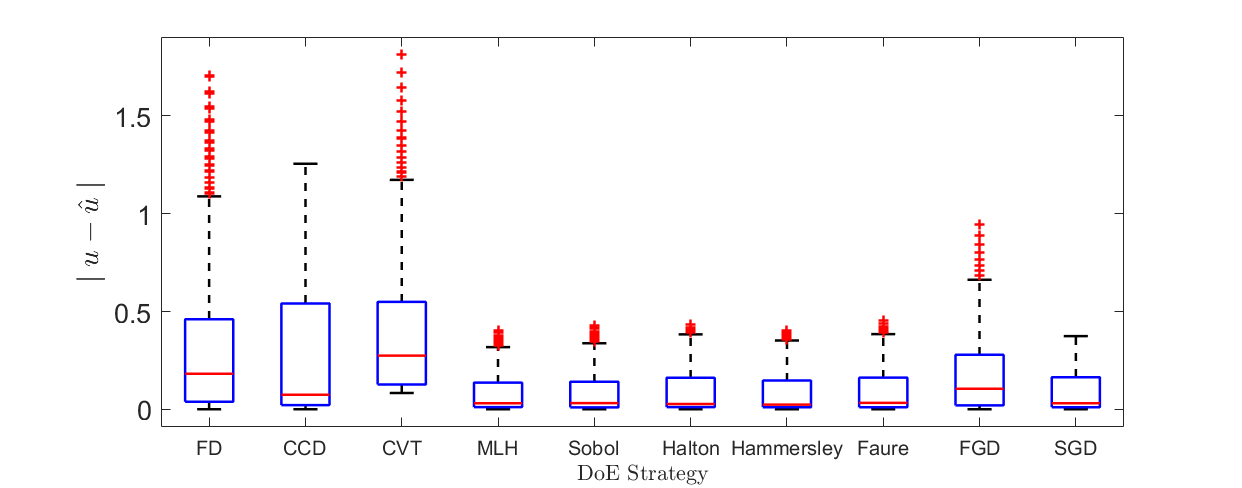}
    \caption{Absolute error between original and predicted responses of Shr\"{o}dinger equation considering different DoE strategies}
    \label{fig:shrodinger_error}
\end{figure*}
With this, a comparative study is performed with all DoE schemes and the absolute errors between actual and predicted responses are shown using Box and Whisker plot as shown in Fig.~\ref{fig:shrodinger_error}. It is seen that FD, CCD and CVT perform the worst by comparing the median values. Apart from FGD, the other DoE strategies such as MLH, SGD, Sobol, Halton, Hammersley, and Faure sampling almost perform the same. Among those, Hammersley sampling slightly performs better. Once the accuracy is established, the predicted responses are evaluated considering the 400 DoE samples generated using Hammersley sampling. Fig.~\ref{fig:shr1} and Fig.~\ref{fig:shr2} show the contour plot of predicted solution of the Shr\"{o}dinger equation and the error between exact and predicted solutions, respectively.

\subsection{Problem 3: 1D - Heat Equation}

\begin{figure}[h!]
    \centering
    \includegraphics[width=0.49\textwidth]{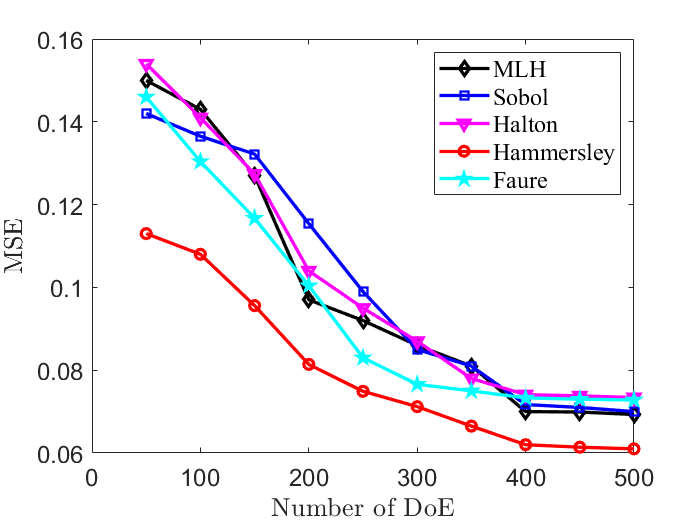}
    \caption{The MSE values for different number of DoE sets for different DoE techniques for heat equation}
    \label{fig:conv_heat}
\end{figure}

\begin{figure*}[h!]
\centering     
\subfigure[]{\label{fig:heat1}\includegraphics[width=0.49\textwidth]{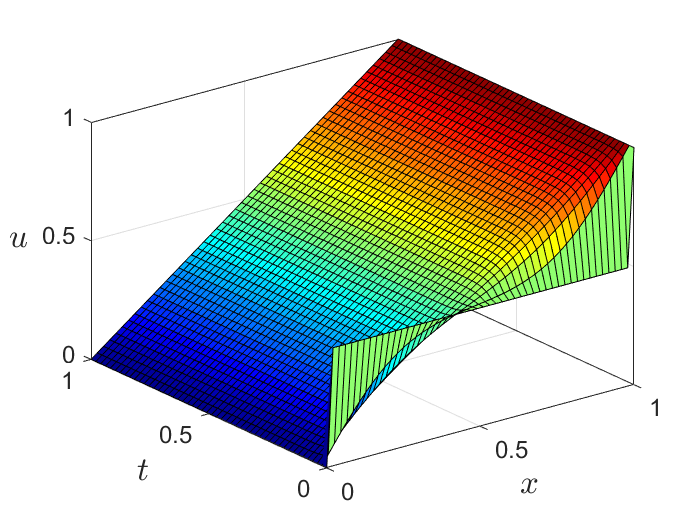}}
\subfigure[]{\label{fig:heat2}\includegraphics[width=0.49\textwidth]{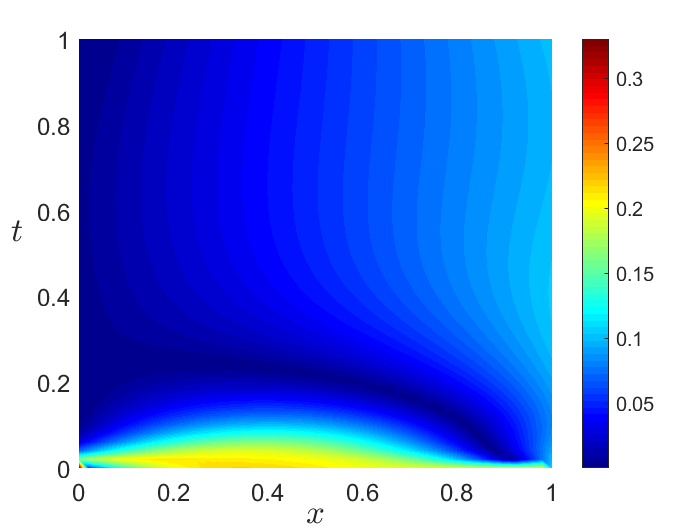}}
\caption{\protect\subref{fig:heat1} Solution for $u$ using PINN and \protect\subref{fig:heat2} Absolute error between exact solution and solution using PINN for heat conduction equation}
\label{fig:heat_cond}
\end{figure*}

As an example, consider the one-dimensional heat equation, which is a second-order PDE that describes heat diffusion through a material and is written as
\begin{equation}\label{eq:heat1}
    \frac{\partial u}{\partial t} = \pdv[2]{u}{x}
\end{equation}
where $(x,t) \in \Xi$ = [0 , $L$] $\times$ [0 ,1]. Initially, the temperature is a nonzero constant. Therefore, the initial condition is given by
\begin{equation}
    u (x , 0) = T_0
\end{equation}
\begin{figure}[h!]
    \centering
    \includegraphics[width=0.52\textwidth]{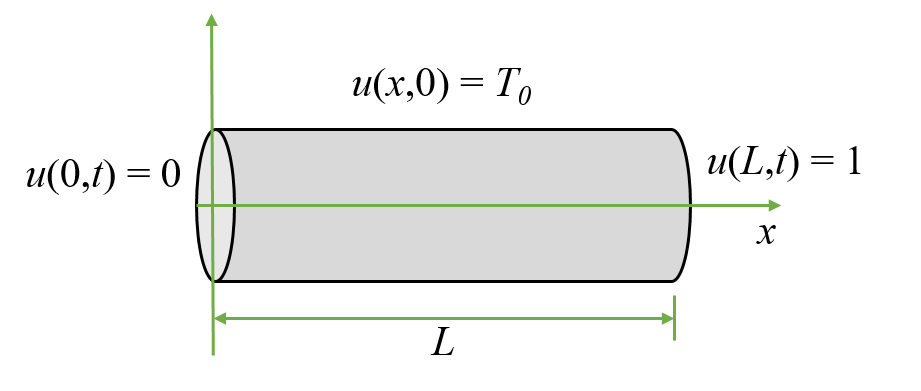}
    \caption{Heat conduction in a rod}
    \label{fig:1dheat}
\end{figure}
Also, the boundary conditions imposed on the system, which are zero at the left boundary and nonzero at the right boundary, is shown in Fig.~\ref{fig:1dheat}. The boundary conditions are expressed as
\begin{equation}
    u (0, t) = 0  \hspace{8mm} u (L ,t) = 1
\end{equation}
In this study, the parameters $L$ and $T_0$ are taken as 1 and 0.5, respectively. The optimal neural network parameters are estimated by minimizing the loss function ${\mathcal{L}} = {\mathcal{L}}_u + {\mathcal{L}}_f$. The two components of the loss function are given by
\begin{subequations}\label{eq:heat12}
\begin{equation}
    {\mathcal{L}}_u = \frac{1}{N_u} \sum_{i=1}^{N_u} \Bigl\vert u (t_u^i,x_u^i) - u^i \Bigl\vert^2
\end{equation}
\begin{equation}
    {\mathcal{L}}_f = \frac{1}{N_f} \sum_{i=1}^{N_f} \Bigl\vert f (t_f^i,x_f^i) \Bigl\vert^2
\end{equation}
\end{subequations}
The loss function corresponding to the initial and boundary conditions is denoted by ${\mathcal{L}}_u$, and the loss function corresponding to the function $f(t,x)$, which is defined as $f = (u_t - u_{xx})$, is denoted by ${\mathcal{L}}_f$. The network contains eight hidden layers, each with 20 neurons. 
 The MSE values for various sizes of DoE sets are shown in Fig.~\ref{fig:conv_heat}. It is seen that 400 samples are needed to stabilize the MSE values. 
\begin{figure*}[h!]
    \centering
    \includegraphics[width=1.0\textwidth]{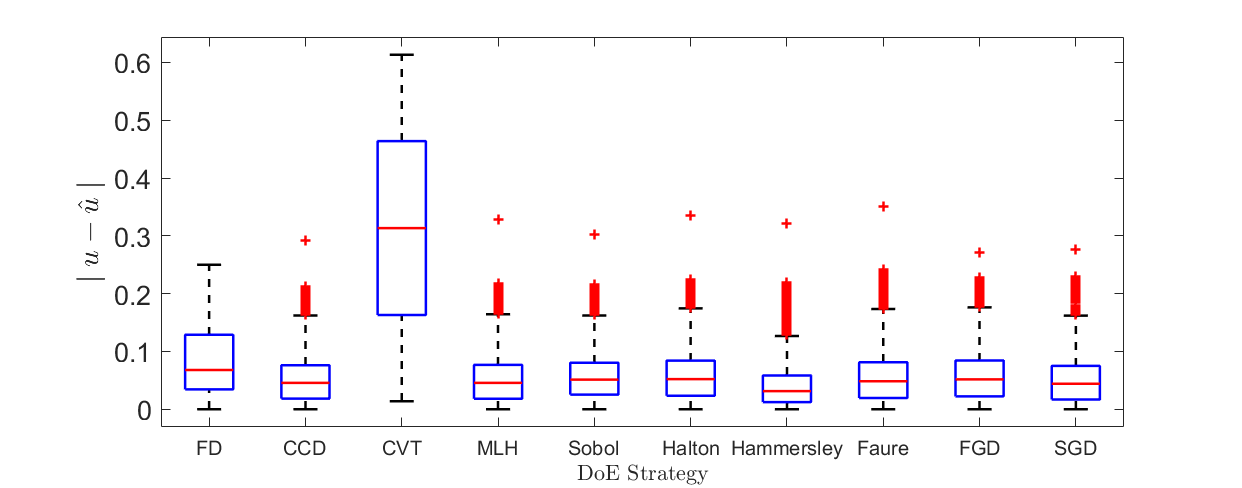}
    \caption{Absolute error between original and predicted responses of heat equation considering different DoE strategies }
    \label{fig:heat_error}
\end{figure*}
With this, a comparative study is performed with all DoE schemes and the absolute errors between actual and predicted responses are shown using a Box and Whisker plot as shown in Fig.~\ref{fig:heat_error}. It is seen that FD and CVT perform the worst by comparing the median values. Among others, Hammersley sampling performs better as its median value is the lowest. Once the accuracy is established, the predicted responses are evaluated considering the 400 DoE samples generated using Hammersley sampling. Fig.~\ref{fig:heat1} and Fig.~\ref{fig:heat2} show the contour plot of the predicted solution of the one-dimensional heat conduction equation and the error between exact and predicted solutions, respectively.

\subsection{Problem 4: Allen - Cahn Equation}

\begin{figure}[h!]
    \centering
    \includegraphics[width=0.49\textwidth]{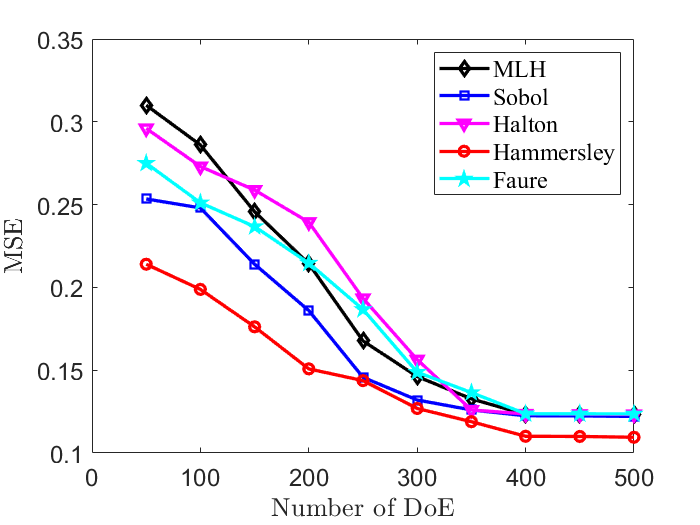}
    \caption{The MSE values for different number of DoE sets for different DoE techniques for Allen-Cahn equation}
    \label{fig:conv_allen}
\end{figure}

\begin{figure*}[h!]
\centering     
\subfigure[]{\label{fig:allen1}\includegraphics[width=0.49\textwidth]{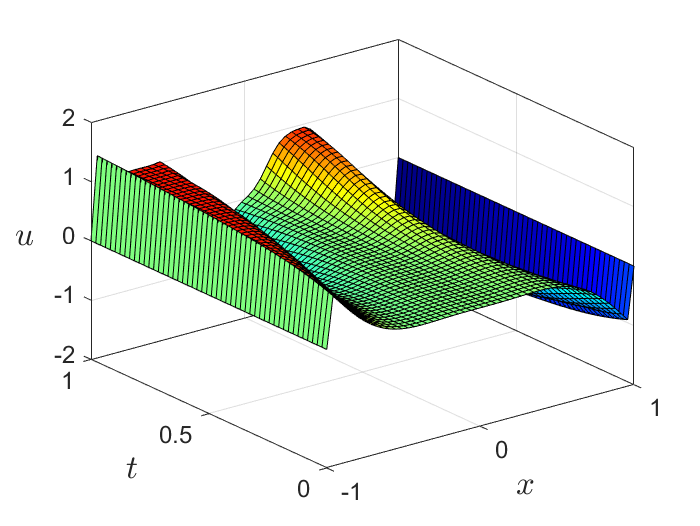}}
\subfigure[]{\label{fig:allen2}\includegraphics[width=0.49\textwidth]{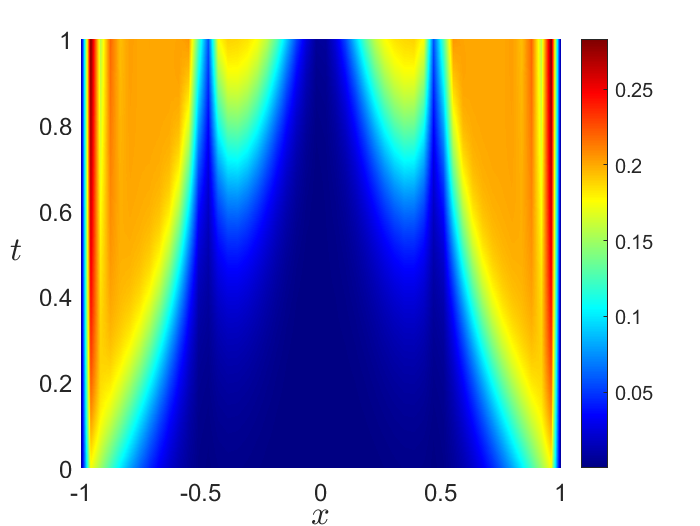}}
\caption{\protect\subref{fig:allen1} Solution for $u$ using PINN and \protect\subref{fig:allen2} Absolute error between exact solution and solution using PINN for reaction - diffusion system represented by Allen-Cahn equation}
\label{fig:allen_cahn}
\end{figure*}

Allen - Cahn equation, a second - order PDE, is used to represent the physical problems such as crystal growth \citep{shah2016efficient}, image segmentation \citep{benevs2004geometrical}, motion by mean curvature flows \citep{shah2018efficient} etc. Mainly, this equation is used to study phase transitions and interfacial dynamics in material science \cite{shah2011numerical}. The governing differential equation is given as 
\begin{equation}\label{eq:ac1}
        \frac{\partial u}{\partial t}  - 0.0001 \pdv[2]{u}{x} + 5 u^3 - 5 u = 0
\end{equation}
where $(x,t) \in \Xi$ = [-1 , 1] $\times$ [0 ,1]. The initial and boundary conditions of the above equation are given by
\begin{equation}
\begin{split}
    u (t = 0, x) &= x^3 \text{cos}(\pi x) \\
    u (t, x = -1) &= u (t, x = 1)\\
      \frac{\partial u}{\partial x} (t, x = -1) &= \frac{\partial u}{\partial x} (t, x = 1)
\end{split}
\end{equation}
The parameters related to the neural networks are estimated by minimizing a loss function, which is expressed in the following form
\begin{equation}
    \mathcal{L} = {\mathcal{L}}_0 + {\mathcal{L}}_b + {\mathcal{L}}_f
\end{equation}
where ${\mathcal{L}}_0$, ${\mathcal{L}}_b$, ${\mathcal{L}}_f$ are the residual loss functions on the initial data, the periodic boundary condition and Allen-Cahn equation, are as follows
\begin{subequations}
\begin{equation}
    {\mathcal{L}}_0 = \frac{1}{N_0} \sum_{j=1}^{N_0}  \Bigl\vert u(0, x_0^j) - u_0^j\Bigl\vert^2 
\end{equation}
\begin{equation}\label{eq:allen_loss2}
\begin{split}
    {\mathcal{L}}_b = \frac{1}{N_b}  \sum_{j=1}^{N_b} \bigg\{&\Bigl\vert u^j (t_b^j, -1) - u^j (t_b^j, 1) \Bigl\vert^2 + \\
    &\Bigl\vert u_x^j (t_b^j, -1) - u_x^j (t_b^j, 1) \Bigl\vert^2 \bigg\}
\end{split}
\end{equation}
\begin{equation}\label{eq:allen_loss3}
    {\mathcal{L}}_f = \frac{1}{N_f} \sum_{j=1}^{N_f} \Bigl\vert f\big(t_f^j, x_f^j\big) \Bigl\vert^2 
\end{equation}
\end{subequations}
In Eq.~\ref{eq:allen_loss3}, $f$ represents the Allen-Cahn equation, which is defined in Eq.~\ref{eq:ac1}, i.e., $f = u_t - 0.0001 u_{xx}+ 5u^3 - 5u$. To solve the second-order PDE representing the Allen-Cahn equation, a neural network is designed which contains four hidden layers, each with 200 neurons. 
The MSE values for various sizes of DoE sets are shown in Fig.~\ref{fig:conv_allen}. It is seen that 400 samples are needed to stabilize the MSE values.
\begin{figure*}[h!]
    \centering
    \includegraphics[width=1.0\textwidth]{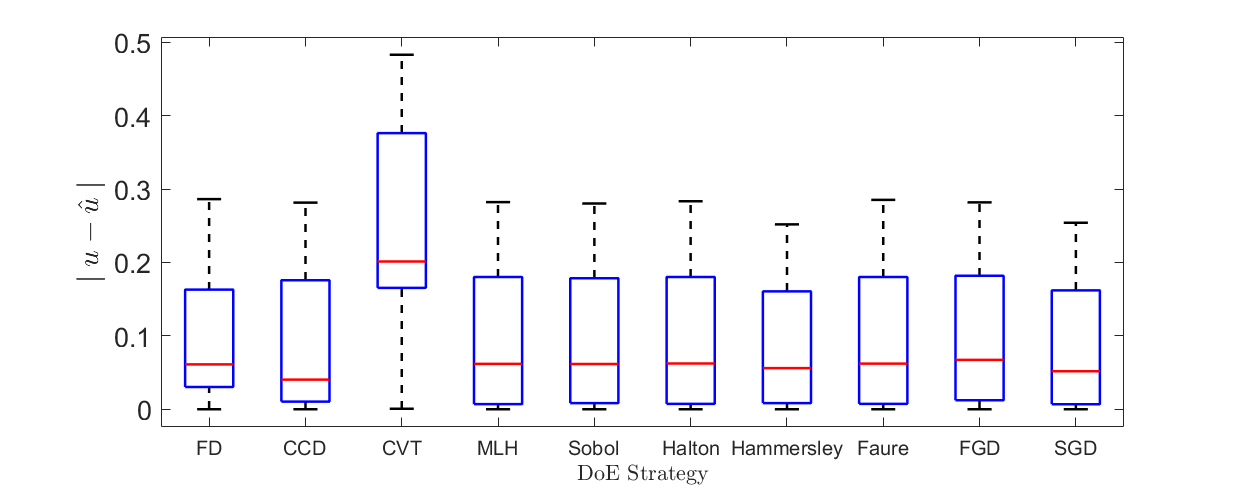}
    \caption{Absolute error between original and predicted responses of Allen-Cahn equation considering different DoE strategies}
    \label{fig:allen_error}
\end{figure*}
Once the size of the DoE set is fixed for quasi-random sampling schemes, a comparative study is performed with all DoE schemes and the absolute errors between actual and predicted responses are shown using the Box and Whisker plot as shown in Fig.~\ref{fig:allen_error}. It is seen that Hammersley sampling performs better as its maximum value of absolute error is the lowest compared to others. Once the accuracy is established, the predicted responses are evaluated considering the 400 DoE samples generated using Hammersley sampling. Fig.~\ref{fig:allen1} and Fig.~\ref{fig:allen2} show the contour plot of the predicted solution of the Allen-Cahn equation and the error between exact and predicted solutions, respectively.

\subsection{Problem 5: Korteweg-de Vries Equation}

\begin{figure}[h!]
\centering     
\includegraphics[width=0.49\textwidth]{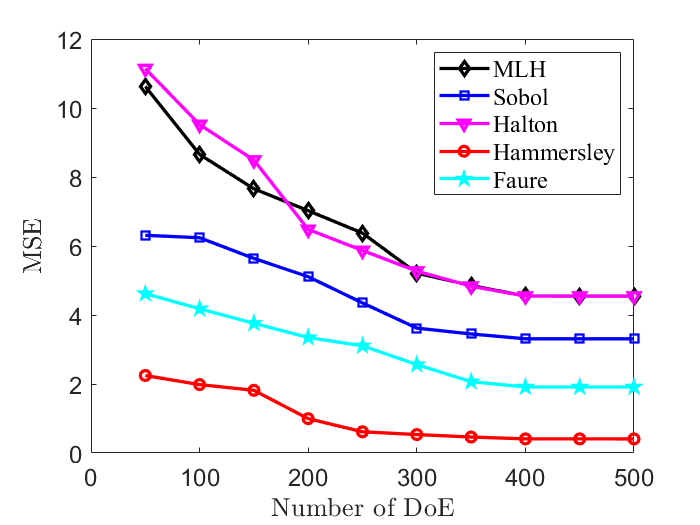}
\caption{The MSE values for different number of DoE sets for different DoE techniques for KdV equation}
\label{fig:conv_kdv}
\end{figure}

As the final example of this study, the Korteweg-de Vries (KdV) equation is considered, which is a nonlinear PDE comprising the higher order derivatives. This equation is used to describe the small amplitude shallow-water waves, ion-phonon waves, and fluctuation phenomena in biological systems, etc. \citep{wu2017inverse,khusnutdinova2018soliton}. The governing KdV equation for shallow-water waves is expressed as
\begin{equation}\label{eq:kdv1}
    \frac{\partial u}{\partial t} + u \frac{\partial u}{\partial x}  + \pdv[3]{u}{x} = 0
\end{equation}
where $(x,t) \in \Xi$ = [0 , 2$\pi$] $\times$ [0 , 1]. With the above equation, periodic boundary conditions, i.e., $u (t, x=0)$ = $u (t, x=2\pi)$ are considered. The initial condition of the KdV equation is given by
\begin{equation}
    u(t = 0, x) = A \sech^2 \bigg(\sqrt{\frac{A}{12}} (x - \pi)\bigg)
\end{equation}
The optimal neural network parameters are determined by minimizing the loss functions (${\mathcal{L}}$ = ${\mathcal{L}}_u$ + ${\mathcal{L}}_f$), which are written as follows
\begin{subequations}\label{eq:kdv12}
\begin{equation}
    {\mathcal{L}}_u = \frac{1}{N_u} \sum_{i=1}^{N_u} \Bigl\vert u (t_u^i,x_u^i) - u^i \Bigl\vert^2
\end{equation}
\begin{equation}
    {\mathcal{L}}_f = \frac{1}{N_f} \sum_{i=1}^{N_f} \Bigl\vert f (t_f^i,x_f^i) \Bigl\vert^2
\end{equation}
\end{subequations}
where ${\mathcal{L}}_u$ and ${\mathcal{L}}_f$ are the loss functions corresponding to the initial and boundary conditions, and the function $f(t,x)$, defined in Eq.~\ref{eq:kdv1}.
The function $f$ in Eq.~\ref{eq:kdv12} represents the KdV equation, defined in Eq.~\ref{eq:kdv1}, i.e., $f = u_t + u u_{x}+ u_{xxx}$. To solve the higher-order PDE representing the KdV equation, a neural network is designed which contains seven hidden layers, each with 120 neurons. 
The MSE values for various sizes of DoE sets are shown in Fig.~\ref{fig:conv_kdv}. It is seen that 350 samples are needed to stabilize the MSE values.

\begin{figure*}[h!]
    \centering
    \includegraphics[width=1.0\textwidth]{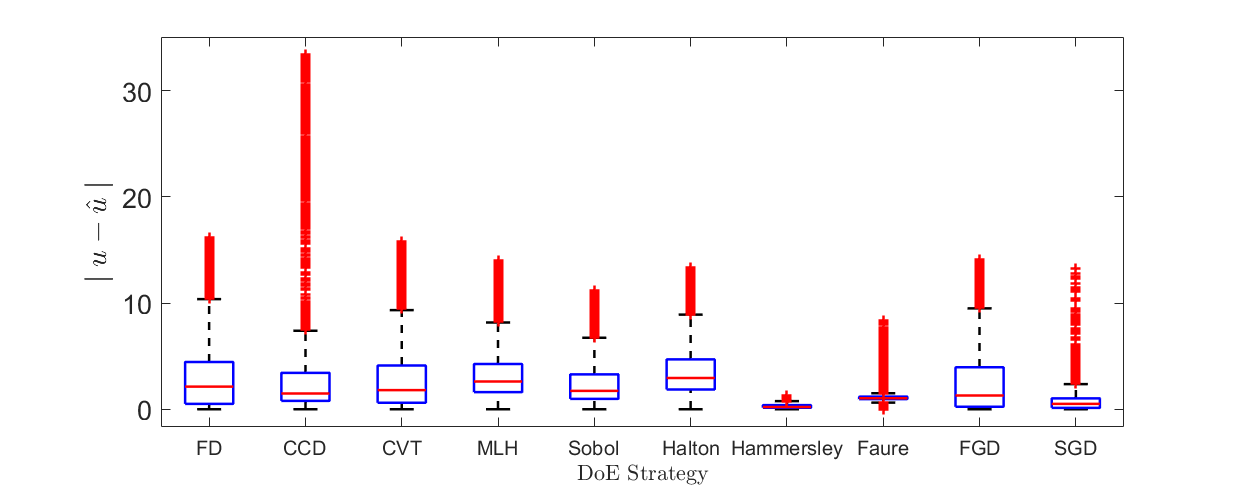}
    \caption{Absolute error between original and predicted responses of KdV equation considering different DoE strategies}
    \label{fig:kdv_error}
\end{figure*}
\begin{figure*}[h!]
\centering     
\subfigure[]{\label{fig:kdv1}\includegraphics[width=0.49\textwidth]{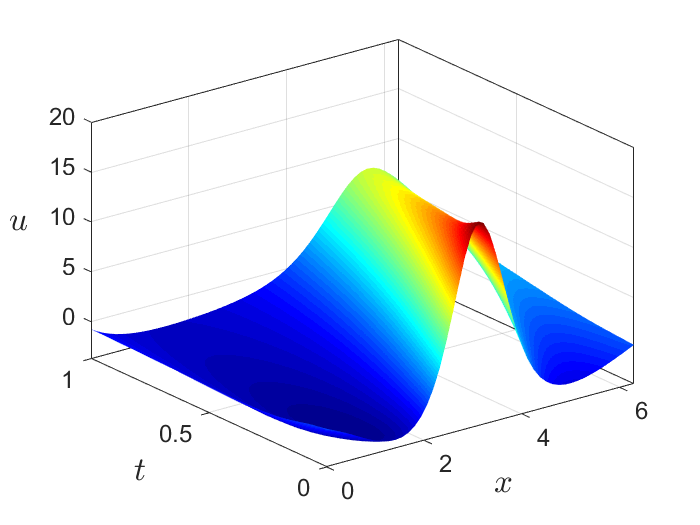}}
\subfigure[]{\label{fig:kdv2}\includegraphics[width=0.49\textwidth]{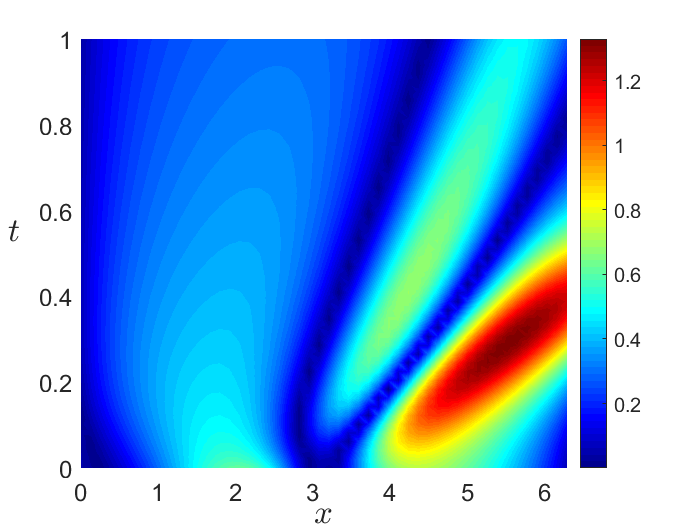}}
\caption{\protect\subref{fig:kdv1} Solution for $u$ using PINN and \protect\subref{fig:kdv2} Absolute error between exact solution and solution using PINN for solitary waves represented by KdV equation}
\label{fig:kdv}
\end{figure*}

Once the size of the DoE set is fixed for quasi-random sampling schemes, a comparative study is performed with all DoE schemes and the absolute errors between actual and predicted responses are shown using the Box and Whisker plot as shown in Fig.~\ref{fig:kdv_error}. It is seen that Hammersley sampling performs better. Once the accuracy is established, the predicted responses are evaluated considering the 350 DoE samples generated using Hammersley sampling. Fig.~\ref{fig:kdv1} and Fig.~\ref{fig:kdv2} show the contour plot of the predicted solution of the KdV equation and the error between exact and predicted solutions, respectively.

\section{Conclusion}\label{sec:conclude}

The selection of the design of experiment strategy plays an important role in predicting the output using surrogate models. The surrogate model predicts the output based on the training data by minimizing the residuals between the output corresponding to the training data and predicted responses. An extensive survey of the different DoE strategies is presented in this study. A review of classical and modern experimental strategy design is also provided. This study is only focused on the physics-informed neural network. In this context, a few salient observations are highlighted below-
\begin{itemize}
    \item Here, five different partial differential equations i.e., Viscous Burger's equation, Shr\"{o}dinger equation, one-dimensional heat equation, Allen-Cahn equation and Korteweg-de Vries equation are considered for investigation purposes. To solve these equations, a physics-based deep neural network method is used, which is trained using the training data generated by different DoE schemes. It is seen that quasi-random sampling techniques, especially the Hammersley sampling scheme, yield better results compared to classical DoE schemes. This argument is established based on comparing the absolute error between the actual and predicted solutions. This reflects that the Hammersley sampling covers the whole design space better than other DoE strategies.
    \item The size of the DoE set affects the accuracy of the surrogate model. For high-fidelity models where the computational budget is expensive, the smallest size of the training set is needed to estimate to reduce the computational budget as well as the accuracy of the prediction. In this study, it is seen that at least 350 to 400 DoE samples are taken to get accurate predicted responses.
\end{itemize}
In general, this paper emphasises the importance of choosing a DoE scheme in order to obtain accurate predicted responses using a physics-informed neural network.In this context, it may be noted that the present study does not consider any sequential method where one sample is added to the DoE set used in the previous iteration based on some performance measures. 

\section*{Acknowledgments}

This study was funded by Natural Sciences and Engineering Research
Council of Canada under the Discovery Grant programs (RGPIN-2019-
05584). Development of the design guideline in this study was supported through the BC Forestry Innovation Investment's (FII) Wood First Program.

\section*{Conflict of interest}
 The authors declare that they have no known competing financial
interests or personal relationships that could have appeared to influence the work reported in this paper.

\bibliography{sn-bibliography}

\end{document}